\documentclass{egpubl}
\usepackage{eg2021}
\usepackage{amsmath}
\usepackage{subfig}
\usepackage{float}
\usepackage{bbding}
\usepackage{pifont}
\usepackage{wasysym}
\usepackage{amssymb}
\usepackage{url}
\usepackage{wrapfig}
\usepackage{threeparttable}

\ConferenceSubmission   
\usepackage[T1]{fontenc}
\usepackage{dfadobe}  

\usepackage{cite}  
\BibtexOrBiblatex
\electronicVersion
\PrintedOrElectronic
\ifpdf \usepackage[pdftex]{graphicx} \pdfcompresslevel=9
\else \usepackage[dvips]{graphicx} \fi

\usepackage{egweblnk} 

\title{Learning a self-supervised tone mapping operator via feature contrast masking loss}

\author[Chao Wang, Bin Chen, Hans-Peter Seidel, Karol Myszkowski, Ana Serrano]
{\parbox{\textwidth}{\centering Chao Wang$^{1}$  Bin Chen$^{1}$ Hans-Peter Seidel$^{1}$ Karol Myszkowskil$^{1}$
        and Ana Serrano$^{2}$
        }
        \\
{\parbox{\textwidth}{\centering $^1$Max-Planck-Institut f\"ur Informatik, Saarbr\"ucken, Germany\\
         $^2$Universidad de Zaragoza, Zaragoza, Spain
       }
}
}

\begin{document}

 \teaser{
  \includegraphics[width=.9\linewidth]{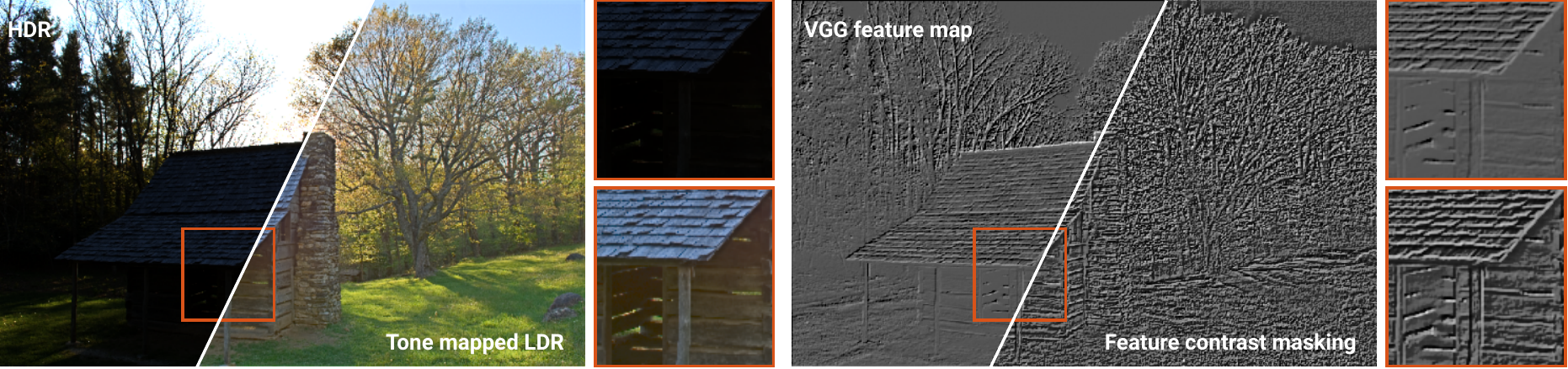}
  \centering
   \caption{
   	Our self-supervised tone mapping operator is directly guided by the input HDR image and a novel feature contrast masking loss that takes into account masking effects present in the Human Visual System. We minimize the difference between the HDR image and the tone mapped image in feature space, after applying our contrast masking model. Left: The input HDR image and our tone mapped result. Right: VGG feature map (1st layer, 12th channel) and our corresponding feature contrast masking response that effectively enhances low contrast image details while compressing high contrasts.}    	  
 \label{fig:teaser}
}

\maketitle
\begin{abstract} 

High Dynamic Range (HDR) content is becoming ubiquitous due to the rapid development of capture technologies. Nevertheless, the dynamic range of common display devices is still limited, therefore tone mapping (TM) remains a key challenge for image visualization. Recent work has demonstrated that neural networks can achieve remarkable performance in this task when compared to traditional methods, however, the quality of the results of these learning-based methods is limited by the training data. Most existing works use as training set a curated selection of best-performing results from existing traditional tone mapping operators (often guided by a quality metric), therefore, the quality of newly generated results is fundamentally limited by the performance of such operators. This quality might be even further limited by the pool of HDR content that is used for training. 
In this work we propose a learning-based self-supervised tone mapping operator that is trained at test time specifically for each HDR image and does not need any data labeling. The key novelty of our approach is a carefully designed loss function built upon fundamental knowledge on contrast perception that allows for directly comparing the content in the HDR and tone mapped images.
We achieve this goal by reformulating classic VGG feature maps into feature contrast maps that normalize local feature differences by their average magnitude in a local neighborhood, allowing our loss to account for contrast masking effects. 
%
We perform extensive ablation studies and exploration of parameters and demonstrate that our solution outperforms existing approaches with a single set of fixed parameters, as confirmed by both objective and subjective metrics.


\end{abstract}  

\section{Introduction}
\label{intro}
High Dynamic Range (HDR) images can reproduce real-world appearance by encoding wide luminance ranges. 
With the fast-paced developments in capturing devices, access to HDR image and video is becoming commonplace. 
Nevertheless, the majority of widespread displays are still limited in the dynamic range they can reproduce, preventing direct reproduction of HDR content. Therefore, the use of tone mapping techniques is yet needed in order to adapt such content to current display capabilities.   

Different tone mapping techniques have been developed for decades~\cite{reinhard2010high,banterle2017advanced}, however the performance of even the most prominent techniques strongly depends on the HDR image content and specific parameter settings~\cite{zhang2019deep, guo2021deep, panetta2021tmo}.
Subjective evaluations of these different techniques indicate that both specific algorithms as well as default parameter settings, as often proposed by their respective authors, do not generalize well across scenes~\cite{ledda2005evaluation,yoshida2005perceptual,vcadik2008evaluation}. In these cases, manual selection of a suitable algorithm and experience in fine-tuning its parameters might be required for high-quality results. This hinders smooth adoption of HDR technology and is the key obstacle in developing machine learning solutions, as we discuss next.
Recently, an increasing number of learning-based tone mapping methods have been proposed that show huge potential in terms of generality and quality with respect to their traditional algorithm-based predecessors. However, there are still some shortcomings. 
First, most learning-based models regard tone mapping as an image restoration task and optimize the network in a fully supervised manner, which requires high-quality paired HDR and tone mapped training data. 
Given a set of HDR scenes as training set, the Tone Mapping Quality Index (TMQI)~\cite{yeganeh2012objective} is typically used to select the best tone mapped image for each scene from a preexisting set of tone mapped results, limiting the quality of newly generated results to that of preexisting methods~\cite{zhang2019deep, rana2019deep, panetta2021tmo}.
And second, since such methods treat tone mapping as an image restoration task instead of an information reduction process, they usually involve large-scale networks. 
Motivated by these observations, in this work we seek an alternative solution that allows for reducing the network size and optimizes information reduction for a given HDR image content.

Since image contrast is arguably one of the key cues in the Human Visual System (HVS) while seeing and interpreting images~\cite{palmer1999vision}, we aim at reproducing perceived contrast in HDR scenes while also ensuring structural fidelity by reproducing visible image details.
To this end, we propose a simple image-specific, self-supervised tone mapping network that is trained at test time and does not require any data labeling.
The only training data is the input HDR image, and the key novelty in our approach is the loss function that directly compares the content in the HDR and tone mapped images.
Since the compared signals present different luminance and contrast ranges, a direct computation of the loss in the feature space, as in e.g., perceptual VGG loss~\cite{simonyan2014very,johnson2016perceptual}, leads to sub-optimal results as we demonstrate in Sec.~\ref{experiment}.
To mitigate this problem, we first propose an adaptive $\mu$-law compression that accounts for the scene brightness and brings HDR image histograms closer to those of low dynamic range images (Fig.~\ref{fig:mu-law}).   
Then, motivated by contrast perception literature~\cite{legge1980contrast,foley1994human,watson1997model,daly2000visual}, we introduce in our loss function a non-linearity considering both the HVS response for stronger contrasts and visual neighborhood masking, and we model it in the network's feature space.
To this end, we first formulate a local contrast measure in the feature space, normalizing local feature differences by their average magnitude in a local neighborhood. This also allows for further abstraction from the magnitude difference between HDR and LDR signals.
Then, we introduce a compressive non-linearity as a function of feature contrast magnitude for the HDR signal, so that for higher magnitudes any departs in feature contrast are less strongly penalized in the loss function.  
This directly translates into compressing higher contrasts while preserving image details in the tone mapped images generated by our network.
Finally, we also introduce feature contrast neighborhood masking, so that the loss function penalizes more weakly changes in feature contrast when similar features are present in the spatial neighborhood of the image.

We perform extensive ablation studies and exploration of parameters and demonstrate that our solution outperforms existing approaches for a single set of loss parameters, as confirmed by both objective and subjective metrics. 
We will make our code publicly available upon acceptance. For more results and explorations please refer to the \href{https://drive.google.com/file/d/1M3Xfnfbk20FML-vaLl1yDn81B_IJMPYh/view?usp=sharing}{ supplementary materials}.
\section{Related work}
\label{related work}

In this section we first summarize related research on contrast perception modeling, and then we discuss tone mapping techniques with emphasis on those that explicitly process contrast as well as more recent learning-based methods.

\subsection{Contrast perception modeling}
\label{related_work_cpm}

Visual sensitivity is affected by a number of key image dimensions such as luminance level~\cite{ahumada1992luminance}, spatial and temporal frequency~\cite{robson1966spatial}, or local image contrast~\cite{watson1989receptive}, as well as their interactions.
In particular, changes in sensitivity as a function of local image contrast are generally termed masking effects. 
There are two main masking effects related to spatial contrast perception that have been widely studied and applied to computer graphics applications: contrast self-masking and visual contrast masking.   
Contrast self-masking~\cite{daly2000visual} is characterized by a strong non-linearity that allows for stronger absolute changes for higher supra-threshold contrasts than for lower near-threshold contrasts before such changes become noticeable~\cite{kingdom1996contrast}.
Visual masking (also called neighborhood masking)~\cite{legge1980contrast,foley1994human,watson1997model,daly2000visual} is a phenomenon in which sensitivity is locally reduced with increases in image local contrast~\cite{daly1992visible}. When contrasts with similar spatial frequencies are present in a close neighborhood, the thresholds for detection of lower contrasts and for change discrimination of higher contrast rise. To model this effect, the input image contrast is decomposed into frequency bands using a filter bank such as a Laplacian pyramid~\cite{peli1990contrast, mantiuk2021fovvideovdp}, a cortex transform~\cite{daly1992visible}, wavelets~\cite{daly2000visual}, or discrete cosine transforms (DCTs)~\cite{watson1993visually}, and then the visual masking is modeled for each frequency band.

Modeling contrast self-masking and neighborhood masking has been proven to be beneficial for several applications including image~\cite{daly2000visual} or video~\cite{Yi2021_CSF} compression, image quality evaluation~\cite{lubin1995visual, mantiuk2021fovvideovdp}, rendering~\cite{bolin1998perceptually, ramasubramanian1999perceptually}, and foveated rendering~\cite{tursun2019luminance}.
Existing works apply such contrast perception models in the primary image contrast domain (or disparity domain~\cite{didyk2010apparent}), and employ predefined filter banks.
Instead, we use a neural network and compute per-channel contrast signals over feature maps, where optimal filters are learned for the task at hand, and we formulate a novel loss function that models contrast self-masking and neighborhood masking in the feature contrast domain.



\subsection{Tone mapping techniques}
We first summarize traditional global and local tone mapping operators 
and then discuss learning-based tone mapping techniques.

\subsubsection{Traditional methods}
Traditional tone mapping methods can be roughly categorized into global methods, that apply the same transfer function to the whole image~\cite{drago2003adaptive,jack1993tone,larson1997visibility,mantiuk2008display}, and local methods, in which the applied function varies for each pixel by taking into account the influence of neighborhood pixels~\cite{fattal2002gradient,reinhard2002photographic,durand2002fast,mantiuk2006perceptual}.
An interested reader may refer to extensive surveys that discuss in length these methods~\cite{reinhard2010high,banterle2017advanced,mantiuk2015ency}.
Other techniques, such as exposure fusion~\cite{mertens2007exposure, raman2009bilateral} directly composite a low dynamic range high-quality image by fusing a set of bracketed exposures, bypassing the reconstruction of an HDR image.
In our pipeline we also fuse several exposures to obtain the final tone mapped image, however, this fusion is performed in feature space instead of in image domain, and our contrast masking loss actively guides this process. 

Recently, a number of model-based tone mapping algorithms achieving higher performance have been proposed. We introduce them here and include them in our comparisons in Sec.~\ref{objective comparisons}.
Shan et al.~\cite{shan2009globally} introduced a method that operates in overlapping windows over the image in which dynamic range compression is optimized globally for the image while window-based local constraints are also satisfied. This allows for both small details and large image structures to be preserved. 
Liang et al.~\cite{liang2018hybrid} designed a hybrid $l_{1}$-$l_{0}$ multi-scale decomposition model that decomposes the image into a base layer, to which an $l_{1}$ sparsity term is imposed to enforce piecewise smoothness, and a detail layer, to which an $l_{0}$ sparsity term is imposed as structural prior, in order to avoid halos and over-enhancement of contrast.  
Li et al.~\cite{li2018clustering} propose to decompose HDR images into color patches and cluster them according to different statistics. Then, for each cluster they apply principal component analysis to find a more compact domain for applying different tone mapping curves.
In general, these traditional methods are model-based and need to introduce prior information, furthermore, they usually require careful parameter tuning which is not user-friendly for non-expert users.

The closest to our goals are gradient domain techniques~\cite{fattal2002gradient,mantiuk2006perceptual,shibata2016gradient} that effectively compress/enhance contrast. Fattal et al.~\cite{fattal2002gradient} consider gradients between neighboring pixels, while Shibata et al.~\cite{shibata2016gradient} employ a base- and detail-layer decomposition, manually pre-selecting parameters to guide contrast manipulations. Mantiuk et al.~\cite{mantiuk2006perceptual} proposes a multi-resolution contrast processing that is driven by a perceptual contrast self-masking model as proposed in~\cite{kingdom1996contrast}. In contrast, our model additionally takes into account neighborhood masking effects, further, contrast masking effects are computed in feature domain instead of traditional image domain.

\subsubsection{Learning-based methods}

Due to the great success of deep learning in image processing tasks, new learning-based tone mapping operators have been proposed during the last years. Most of these works fall under the category of supervised methods, i.e., they need HDR-LDR (low dynamic range) image pairs in order to train their proposed models. Patel et al.~\cite{patel2017generative} propose to train a generative adversarial network (GAN) in order to learn a combination of traditional tone mapping operators that allows for better generalization across scenes. During training, in order to select the target tone mapped image, they select the best scoring tone mapped image (using the TMQI metric~\cite{yeganeh2012objective}) among a set of tone mapped results using different traditional methods. Rana et al.~\cite{rana2019deep} instead propose using a multi-scale conditional generative adversarial network, and then followed the same procedure for selecting tone mapped images for training. Zhang et al.~\cite{zhang2019deep} use a carefully designed loss function to push tone mapped images into the natural image manifold. The target tone mapped images for training are manually adjusted by photographers using the tone mapping operators available in Photomatix2 and HDRToolbox. Su et al.~\cite{su2021explorable} propose an explorable tone mapping network based on BicycleGAN~\cite{zhu2017multimodal} and use LuminanceHDR to generate suitable tone mapped target images, selecting the top-scoring ones using the TMQI metric. To mitigate the challenge of finding target tone mapped images suitable for training, Panetta et al.~\cite{panetta2021tmo} use low-light images based on the insight that they have under-exposed regions that model well the distribution of HDR images while also having characteristics considered as under-exposed when viewed in displays with limited dynamic range. The method proposed by Yang et al.~\cite{yang2018image} allows to recover image details by training an end-to-end network for reconstructing HDR images from LDR ones, and then performing tone mapping. They use Adobe Photoshop as a black box to empirically generate ground truth tone mapped images with human supervision. 
Recently, Zhang et al.~\cite{9454333} propose a semi-supervised method by combining unsupervised losses and a supervised loss. In this manner, their method only requires a few HDR-LDR pairs with well tone mapped images. For the supervised loss term, they use LDR images from the previously discussed work of Zhang et al.~\cite{zhang2019deep}. Inspired by image quality assessment metrics, Guo and Jiang~\cite{guo2021deep} also propose a semi-supervised method and obtain HDR-LDR pairs from fine-tuning raw bracketed exposures using Adobe Photoshop. 
%

These learning methods are intrinsically limited by the input data they see during training. Therefore, using images tone mapped with existing methods as target, although allows for training new models with better generalization, fundamentally limits the quality that such new learned models can achieve. In contrast, we propose a self-supervised network that only takes as input the original HDR image for training, and learns a tone mapping operator relying on a carefully designed loss function that takes into account contrast masking effects present in the Human Visual System. 
To our knowledge, the only work that does not need carefully selected HDR-LDR image pairs is the work of Hou et al.~\cite{hou2017deep}, however, they rely on combining feature losses from different layers chosen empirically and only demonstrate their approach for two selected images.

\section{Proposed Method}
\label{proposed methods}

\begin{figure*}[thb]
	\centering
	\includegraphics[width=\linewidth]{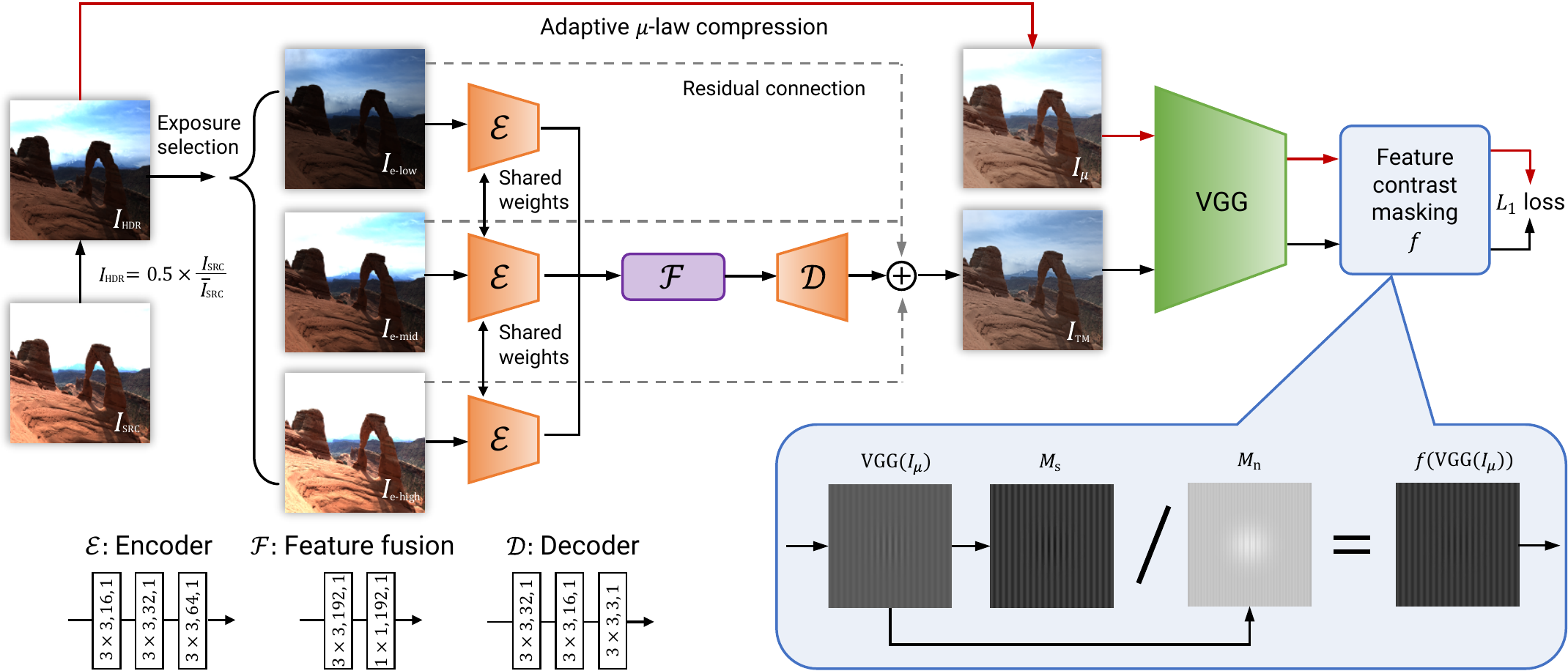}
	\hfill
	\caption{\label{fig:pipeline}%
		Overview of our method. The input HDR image $I_{\scalebox{.9}{$\scriptscriptstyle \textrm{SRC}$}}$ is first normalized into $I_{\scalebox{.9}{$\scriptscriptstyle \textrm{HDR}$}}$, and then three exposures $I_{\scalebox{.99}{$\scriptscriptstyle \textrm{e-low}$}}$, $I_{\scalebox{.99}{$\scriptscriptstyle \textrm{e-mid}$}}$, and $I_{\scalebox{.99}{$\scriptscriptstyle \textrm{e-high}$}}$ are selected. 
		A network with learnable HDR-image-specific weights is used to encode ($\mathcal{E}$) each exposure into feature space, fuse ($\mathcal{F}$) the resulting features, and decode ($\mathcal{D}$) them into the output tone mapped image. 
		The appearance and contrast of the output tone mapped image $I_{\scalebox{.9}{$\scriptscriptstyle \textrm{TM}$}}$ is directly guided by the normalized HDR $I_{\scalebox{.9}{$\scriptscriptstyle \textrm{HDR}$}}$ that is further processed into $I_{\mu}$ using an adaptive $\mu$-law compression. 
		Both $I_{\scalebox{.9}{$\scriptscriptstyle \textrm{TM}$}}$ and $I_{\mu}$ are transformed into feature space (VGG) and compared taking into account our novel contrast masking model $f$.
		As shown in the bottom-right inset $I_{\mu}$ is transformed into the feature space $\mathrm{VGG}(I_{\mu})$, where the ratio between feature contrast self-masking $M_\mathrm{s}$ and feature contrast neighborhood masking $M_n$ define the feature contrast masking model $f(\mathrm{VGG}(I_{\mu}))$. 
		Finally, the $L_1$ loss is computed between  $f(\mathrm{VGG}(I_{\mu}))$ and $f(\mathrm{VGG}(I_{\scalebox{.9}{$\scriptscriptstyle \textrm{TM}$}}))$. 
		The inset in the bottom-left shows the structure of decoder ($\mathcal{E}$), feature fusion ($\mathcal{F}$), and decoder ($\mathcal{D}$) networks.
	}
\end{figure*}

In this section we present our image-specific, self-supervised tone mapping network, whose structure is shown in Fig.~\ref{fig:pipeline}.
The input HDR image is first normalized, and then decomposed into three differently exposed LDR images  (Section~\ref{mef}).  
The three exposures are first transformed into feature space by an encoder with shared weights, then they are fused in this feature space, and finally decoded into the output tone mapped image (Section \ref{network structure}).
To compute the training loss, first the normalized HDR image is processed by an adaptive $\mu$-law compression that brings its distribution closer to typical LDR image histogram distribution (Section~\ref{mu law}).   
Then, the processed HDR image along with its tone mapped counterpart go through a VGG network to derive their respective feature spaces (we employ VGG19 \cite{simonyan2014very}, which we denote VGG for brevity). 
Finally, we compute the L1 loss in feature space, however, instead of the standard perceptual loss between corresponding features~\cite{simonyan2014very,johnson2016perceptual}, we compute feature contrast and model contrast self-masking and neighborhood masking effects (Section~\ref{vml}). 

\subsection{Multiple Exposure Selection}
\label{mef}
HDR images are stored in linear intensity space and might feature extremely large dynamic range compared to LDR images.
Moreover, the distribution of pixel intensities in the HDR image is also unbalanced, where pixels with very high intensity have large values but correspond small image regions, while low-brightness pixels usually cover larger portions of the image~\cite{eilertsen2017hdr,panetta2021tmo}. 
In neural network applications, to align such HDR image characteristics to those of LDR images, the logarithm of HDR pixel intensities is often applied~\cite{eilertsen2017hdr,zhang2019deep, su2021explorable}.
This way a compressive response of the HVS to increasing luminance values (the Weber-Fechner law) is modeled as it is common in the tone mapping literature~\cite{reinhard2010high}.
However, as we detail in Section~\ref{ablation studies}, we found that our tone mapping network leads to better results when multiple differently exposed images with linear pixel intensity relation are used instead.

As HDR images are typically stored as relative positive values, before choosing the exposures, we first derive the normalized $I_{\scalebox{.9}{$\scriptscriptstyle \textrm{HDR}$}}$ image, where each pixel value is multiplied by 0.5, and divided by the mean of the original HDR image $I_{\scalebox{.9}{$\scriptscriptstyle \textrm{SRC}$}}$ intensity~\cite{endoSA2017}.
To estimate the exposure range for each exposure we employ an automatic procedure originally proposed for HDR image quality evaluation~\cite{andersson2021visualizing}.
This way we obtain the low $e_{\mathrm{low}}$ and high $e_{\mathrm{high}}$ exposures\footnote{Note that the goal of Andersson et al.~\cite{andersson2021visualizing} is to get aesthetical results. We divide by two both $e_{\mathrm{low}}$ and $e_{\mathrm{high}}$, resulting in less dark and less bright exposures, respectively. Our goal is that all relevant content is within reasonable pixel values (not too dark, not too bright).}
, and additionally we derive an intermediate third exposure as $e_{\mathrm{mid}} = (e_{\mathrm{low}} + e_{\mathrm{high}})/2$. 
As we show in detail in the supplemental, we found that selected this way three exposures lead to overall good results. Only two exposures are not sufficient and four or more exposures do not improve tone mapping quality, while increasing the computation cost.
Note that in contrast to standard multi-exposure 8-bit LDR images, our three exposure selection is sufficient to represent high dynamic range information for tone mapping purposes. We do not perform pixel quantization, so the only information loss is due to clipping higher intensities into the range [0,1] by the $\mathrm{clip()}$ function:
\begin{equation}
	\label{eq:mepB}
I_{\scalebox{.99}{$\scriptscriptstyle \textrm{e-x}$}} = \mathrm{clip}({2}^{e_{\mathrm{x}}} I_{\scalebox{.9}{$\scriptscriptstyle \textrm{HDR}$}}),
\end{equation}
\noindent
where 
$I_{\scalebox{.99}{$\scriptscriptstyle \textrm{e-x}$}}$ represents one of multi-exposure images with the exposure factor $e_{\mathrm{x}}$.  
\subsection{Tone Mapping Network}
\label{network structure}

The tone mapping network is composed of an encoder $\mathcal{E}$, a feature fusion module $\mathcal{F}$, and a decoder $\mathcal{D}$.
All details on the number of layers, as well as the per-layer kernel size, channel number, and stride extent are specified in the bottom-left corner of Fig.~\ref{fig:pipeline}. All layers use Relu \cite{nair2010rectified} as the activation function except for the last layer using sigmoid.
The three input exposures 
are used as an input to the encoder network $\mathcal{E}$ with shared weights between the exposures.
The resulting feature maps are then processed by the fusion module $\mathcal{F}$ that is essential for information exchange between the exposures.
Our loss function, guided by the HDR image, ensures that the contrast of the tone mapped image is reproduced from the most meaningful exposure, while its magnitude is mediated through complementary information from other exposures.
Saturated (clipped) or poorly exposed regions in any exposure are strongly penalized by the loss, since they differ from the input HDR image. 
Image-specific learned filter weights lead to an optimized HDR image representation. 
Finally, the decoder $\mathcal{D}$ reconstructs the output tone mapped image $I_{\scalebox{.9}{$\scriptscriptstyle \textrm{TM}$}}$.
For more stable training and faster convergence, residual connections are added between each input and the output~\cite{he2016deep}.

\subsection{Adaptive $\mu$-law compression}
\label{mu law}
\begin{figure}[thb]
	\centering
	\includegraphics[width=\linewidth]{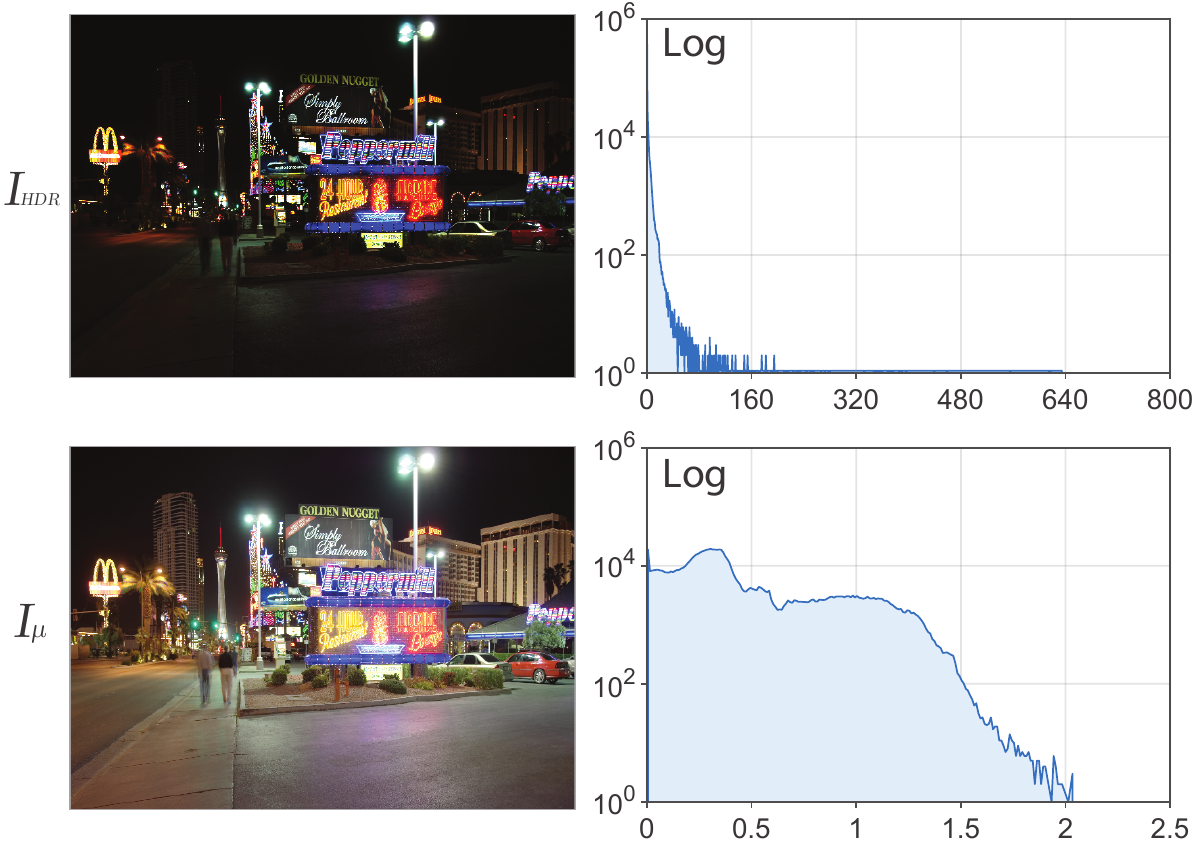}
	\hfill
	\caption{\label{fig:mu-law}%
		Image appearance and logarithmic histogram of pixel intensities for a transformed HDR image. Top Row:  $I_{\scalebox{.9}{$\scriptscriptstyle \textrm{HDR}$}}$ with the mean normalized to 0.5. Bottom row: $I_\mathrm{\mu}$ derived using the adaptive $\mu$-law compression (Eqs.~\ref{eq:hmu} and \ref{eq:fitmu}). Notice the dramatically different scales on the $x$-axis.
}
\end{figure}

Our image-specific tone mapping network is self-supervised by the input HDR image $I_{\scalebox{.9}{$\scriptscriptstyle \textrm{SRC}$}}$ that requires its transformation by the VGG network into a feature space (Fig.~\ref{fig:pipeline}).
We first adapt the range of intensity values by converting  $I_{\scalebox{.9}{$\scriptscriptstyle \textrm{SRC}$}}$ into $I_{\scalebox{.9}{$\scriptscriptstyle \textrm{HDR}$}}$ as discussed in Section~\ref{mef}.
However, as shown in Fig.~\ref{fig:mu-law} (top row) this image $I_{\scalebox{.9}{$\scriptscriptstyle \textrm{HDR}$}}$ still has a strong skew of its histogram towards low intensity values (typical for HDR images~\cite{eilertsen2017hdr,panetta2021tmo}), which strongly differs from LDR image characteristics. Since LDR images are typically used for training VGG, we resort to a $\mu$-law compression to correct for these problems:
\begin{equation}
	I_{\mu} = \frac{\log(1 + \mu I_{\scalebox{.9}{$\scriptscriptstyle \textrm{HDR}$}})}{ \log(1 + \mu)},
	\label{eq:hmu}
\end{equation}

\begin{figure}[thb]
	\centering
	\includegraphics[width=\linewidth]{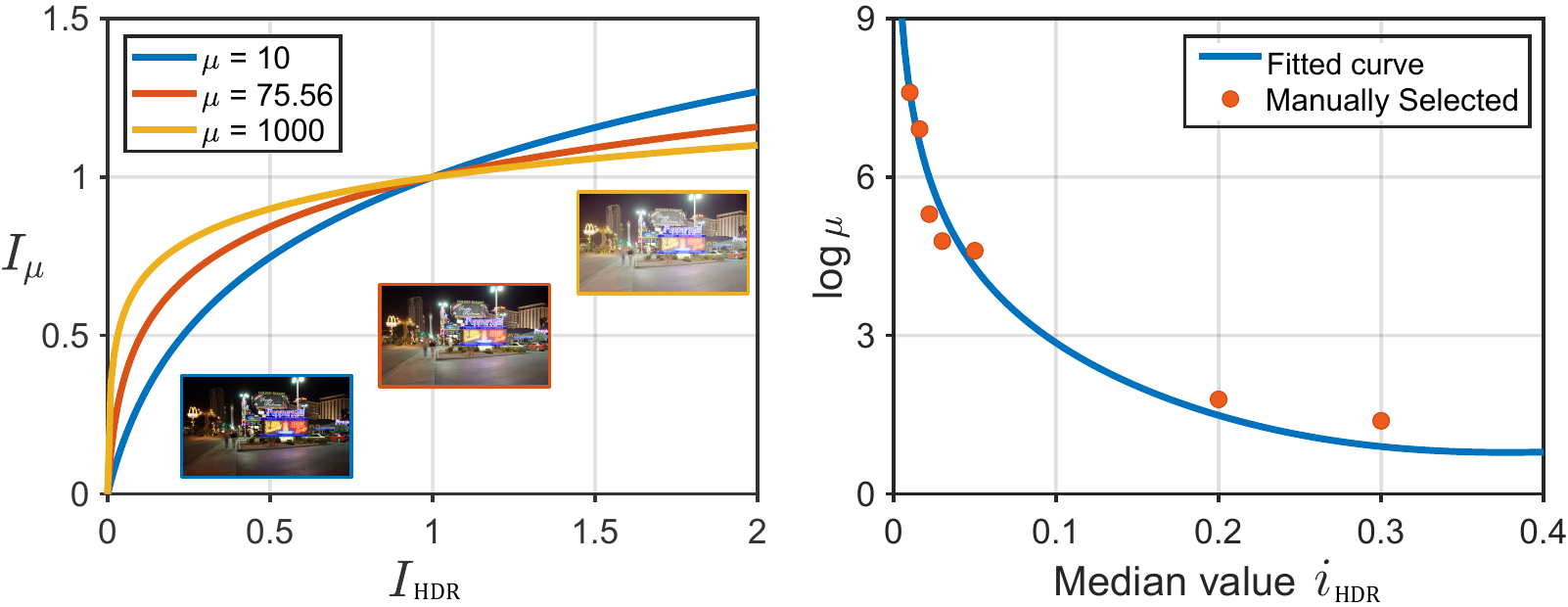}
    \caption{\label{fig:mu_curve}%
		Left: Adaptive $\mu$-law compression (Eq.~\ref{eq:hmu}) for different $\mu$ values and the corresponding compressed HDR images. In this case $\mu=75.56$ has been selected for this image using Eq.~\ref{eq:fitmu}.  Right: Experimentally derived $\mu$ selection as a function of HDR image median intensity $i_{\scalebox{.9}{$\scriptscriptstyle \textrm{HDR}$}}$.}
\end{figure}

This algorithm is widely used in HDR image coding~\cite{jinno2011mu} and inverse tone mapping~\cite{wu2018hdr,Marcel:2020:LDRHDR} to re-arrange the intensity distribution.
Fig.~\ref{fig:mu-law} (bottom row) shows an example of this transformation, where image details become more visible and the long tail in the histogram is corrected.
Typically a fixed scaling constant $\mu$ is used to derive the transformed image $I_{\mu}$, but as we show in the insets in Fig.~\ref{fig:mu_curve} (left), the resulting image appearance strongly depends on the selected $\mu$ value.
We observe that the choice of $\mu$ also affects greatly the performance of the VGG-based loss that drives our tone mapping network, where larger $\mu$ values are required for darker HDR images. 
We propose an adaptive $\mu$-law compression, where the $\mu$ value changes with the median pixel intensity value $i_{\scalebox{.9}{$\scriptscriptstyle \textrm{HDR}$}}$ that is computed for the $I_{\scalebox{.9}{$\scriptscriptstyle \textrm{HDR}$}}$ image:
\begin{equation}
	\mu = \lambda_1 (i_{\scalebox{.9}{$\scriptscriptstyle \textrm{HDR}$}})^{\gamma_1}  + \lambda_2 (i_{\scalebox{.9}{$\scriptscriptstyle \textrm{HDR}$}})^{\gamma_2},
	\label{eq:fitmu}
\end{equation}
\noindent
with fitted parameter values $\lambda_1 = 8.759$, $\gamma_1 = 2.148$, $\lambda_2 = 0.1494$, and $\gamma_2 = -2.067$.
We derive this function experimentally for a number of representative HDR images featuring different appearances as well as different $i_{\scalebox{.9}{$\scriptscriptstyle \textrm{HDR}$}}$ (i.e., brightness).
We use the TMQI quality metric \cite{yeganeh2012objective} to select the best performing $\mu$ values, and then by visual inspection we confirm this selection.
Fig.~\ref{fig:mu_curve} (right) shows the fitted curve based on this procedure.

\subsection{Feature Contrast Masking Loss}
\label{vml}

\begin{figure*}[tb]
	\centering
	\includegraphics[width=\linewidth]{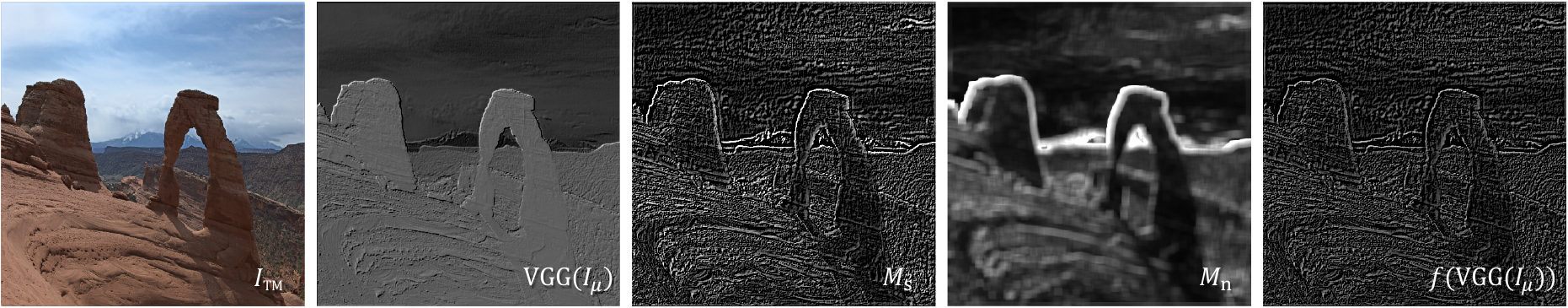}
	\hfill
	\caption{\label{fig:feature_masking_demo}%
		Feature contrast masking visualization of VGG feature maps (1st layer, 18th channel). From left to right: tone mapped image $I_{\scalebox{.9}{$\scriptscriptstyle \textrm{TM}$}}$, original VGG feature map $\mathrm{VGG}(I_{\mu})$, and feature contrast self-masking $M_\mathrm{s}$, neighborhood masking $M_\mathrm{n}$, and final $f(\mathrm{VGG}(I_{\mu}))$ masking terms.
	}
\end{figure*}

In this section we propose the feature contrast masking (FCM) loss that guides our tone mapping network to reproduce image details and overall perceived contrast. 
To this end, we first model feature contrast, and then introduce self contrast masking and neighborhood masking for such feature contrast inspired by their analogues in image domain described in Sec. \ref{related_work_cpm}.

\paragraph*{Feature contrast}

While the HDR image representation $I_{\mu}$, which results from the adaptive $\mu$-law compression (Sec.~\ref{mu law}), greatly facilitates its meaningful processing by the VGG network, still significant intensity differences might exist with respect to its tone mapped version $I_{\scalebox{.9}{$\scriptscriptstyle \textrm{TM}$}}$.
Such intensity differences translate into corresponding differences in the feature magnitude when $I_{\mu}$ and $I_{\scalebox{.9}{$\scriptscriptstyle \textrm{TM}$}}$ are transformed by the VGG network (Fig.~\ref{fig:pipeline}).
To further reduce such feature magnitude differences we compute per-channel a local feature contrast:
\begin{equation}
	C_p = \frac{f_\mathrm{p} - \bar{f}_\mathrm{p}}{\lvert\bar{f}_\mathrm{p}\rvert+\epsilon},
	\label{eq:featcontr}
\end{equation} 
\noindent
where $f_\mathrm{p}$ denotes the feature magnitude at pixel $p$, $\bar{f}_\mathrm{p}$  is the Gaussian filtered feature value computed for the patch $\mathcal{P}$ that is centered at $p$, and $\epsilon$ is a small constant to avoid division by zero.
We experimentally set the patch $\mathcal{P}$ size to 13$\times$13 pixels (refer to the supplemental for details). 
Effectively, in the numerator the feature difference with respect to its local neighborhood is first computed and then normalized.
Such normalization enables further reduction of the impact of differences in absolute feature magnitudes between $I_{\mu}$ and $I_{\scalebox{.9}{$\scriptscriptstyle \textrm{TM}$}}$.
Note that Eq.~\ref{eq:featcontr} is aligned with contrast definitions for images that also use Laplacian and Gaussian filter responses in the numerator and denominator, respectively~\cite{peli1990contrast}.  
More complex filter banks such as Laplacian pyramid~\cite{mantiuk2021fovvideovdp}, cortex transform~\cite{daly1992visible}, wavelets~\cite{daly2000visual}, or discrete cosine transforms (DCT)~\cite{watson1993visually} are often used for advanced contrast processing operations.
 
\paragraph*{Feature contrast self-masking}

An important property of contrast perception is a higher sensitivity to contrast discrimination for lower contrasts rather than suprathreshold contrasts~\cite{kingdom1996contrast}, which is often called contrast self-masking \cite{daly2000visual}.
In the context of tone mapping this means that even small contrast changes can be perceived in low-contrast image regions, which are often neglected when global tone mapping operators are applied~\cite{mantiuk2015ency}.
Conversely, for larger image contrast, even strong contrast compression might remain invisible. Many tone mapping operators take advantage of this effect~\cite{durand2002fast,fattal2002gradient,mantiuk2006perceptual}.
Different from conventional methods, we do not model contrast self-masking in the image contrast domain, but rather for feature contrast domain $C$ (we skip the pixel index $p$ for brevity) as defined in Eq.~\ref{eq:featcontr}:
\begin{equation}
	M_\mathrm{s} = \mathrm{sign}(C)\lvert C\rvert ^{\alpha}, 
\end{equation} 
\noindent
where $M_\mathrm{s}$ denotes a non-linear response to feature contrast magnitude controlled by the compressive power factor $\alpha$. The function $\mathrm{sign}(x) = \frac{x}{\lvert x\rvert + \epsilon}$ preserves the feature contrast polarity in $M_\mathrm{s}$.
We experimentally set $\alpha_{I_{\mu}}=0.5$ when processing $I_{\mu}$ while we keep $\alpha_{I_{\scalebox{.9}{$\scriptscriptstyle \textrm{TM}$}}}=1.0$ for $I_{\scalebox{.9}{$\scriptscriptstyle \textrm{TM}$}}$, allowing for visually pleasant low-contrast detail enhancement (refer to the supplemental for more details).
Fig.~\ref{fig:feature_masking_demo} visualizes a feature map $\mathrm{VGG}(I_{\mu})$ for a selected VGG channel, as well as our response $M_\mathrm{s}$ to feature contrast $C$. 
As can be seen $M_\mathrm{s}$ vividly responds for low-contrast details in the sky and rocks that are strongly suppressed in $\mathrm{VGG}(I_{\mu})$.
Note that when $\mathrm{VGG}(I_{\mu})$ is directly used in the perceptual VGG loss 
$\mathcal{L}_{\scalebox{.9}{$\scriptscriptstyle \textrm{VGG}$}} = \left |\left| \mathrm{VGG}(I_\mu) - \mathrm{VGG}(I_{\scalebox{.9}{$\scriptscriptstyle \textrm{TM}$}})\right | \right|_{1}$ 
\cite{johnson2016perceptual} that will drive the tone mapping operation, such details are likely to be neglected in the resulting $I_{\scalebox{.9}{$\scriptscriptstyle \textrm{TM}$}}$ due to low penalty in the loss.

\paragraph*{Feature contrast neighborhood masking}

Inspired by successful applications of neighborhood masking, 
as discussed in Sec.~\ref{related_work_cpm}, we also model feature contrast neighborhood masking.
Image contrast neighborhood masking is performed selectively for different spatial frequency bands that requires image decomposition by a filter bank \cite{daly2000visual,lubin1995visual}.
We approximate this process by modeling feature contrast neighborhood masking per channel, where features with similar frequency characteristics are naturally isolated.  
Our goal is to suppress the magnitude of $M_\mathrm{s}$ when there is a high variation of feature magnitudes $f_p$ in the local neighboring of pixel $p$ that we measure as:
\begin{equation}
	M_\mathrm{n} = \frac{\sigma_\mathrm{b}}{\lvert\mu_\mathrm{b}\rvert+\epsilon},
\end{equation} 
\noindent
where $\mu_\mathrm{b}$ and  $\sigma_\mathrm{b}$ denote the mean and standard deviation of feature magnitude $f_p$ in the patch $\mathcal{P}$ that is centered at pixel $p$.
Again, we experimentally set the patch $\mathcal{P}$ size to 13$\times$13 pixels. 
Finally, our feature contrast masking is calculated as the ratio of self and neighborhood masking:
\begin{equation}
	f(\mathrm{VGG}(I)) =\frac{M_\mathrm{s}}{1+M_\mathrm{n}}
	\label{eq:masking}
\end{equation}
As can be seen in Fig.~\ref{fig:feature_masking_demo}, $M_\mathrm{n}$ vividly responds in the regions with high local feature variation as seen in $\mathrm{VGG}(I_{\mu})$, so that the final feature contrast masking measure $f(\mathrm{VGG}(I_{\mu}))$ is strongly suppressed in such regions.  
In particular, this means that in the regions of strong image contrast, such as the horizon line, $f(\mathrm{VGG}(I_{\mu}))$ is relatively much smaller with respect to the orginal $\mathrm{VGG}(I_{\mu})$ feature magnitudes.
Consequently, when including $f(\mathrm{VGG}(I_{\mu}))$ into the loss computation that drives tone mapping, the penalty for any distortion of such high contrast is much smaller than in the perceptual VGG loss $\mathcal{L}_{\scalebox{.9}{$\scriptscriptstyle \textrm{VGG}$}}$ that directly employs $\mathrm{VGG}(I_{\mu})$.
Effectively, this gives the tone mapping network more freedom for compressing image contrast in such regions. 

We compute our feature contrast masking (FCM) loss $\mathcal{L}_{\scalebox{.9}{$\scriptscriptstyle \textrm{FCM}$}}$
as the $L_1$ loss between the masked feature maps $f()$ for the transformed input HDR image $I_\mu$ and the output tone mapped image $I_{\scalebox{.9}{$\scriptscriptstyle \textrm{TM}$}}$:
\begin{equation}
	\mathcal{L}_{\scalebox{.9}{$\scriptscriptstyle \textrm{FCM}$}} = \left |\left| f(\mathrm{VGG}(I_\mu)) - f(\mathrm{VGG}(I_{\scalebox{.9}{$\scriptscriptstyle \textrm{TM}$}}))\right | \right|_{1}
	\label{eq:loss}
\end{equation}

\begin{figure}[bth]
	\centering
	\includegraphics[width=\linewidth]{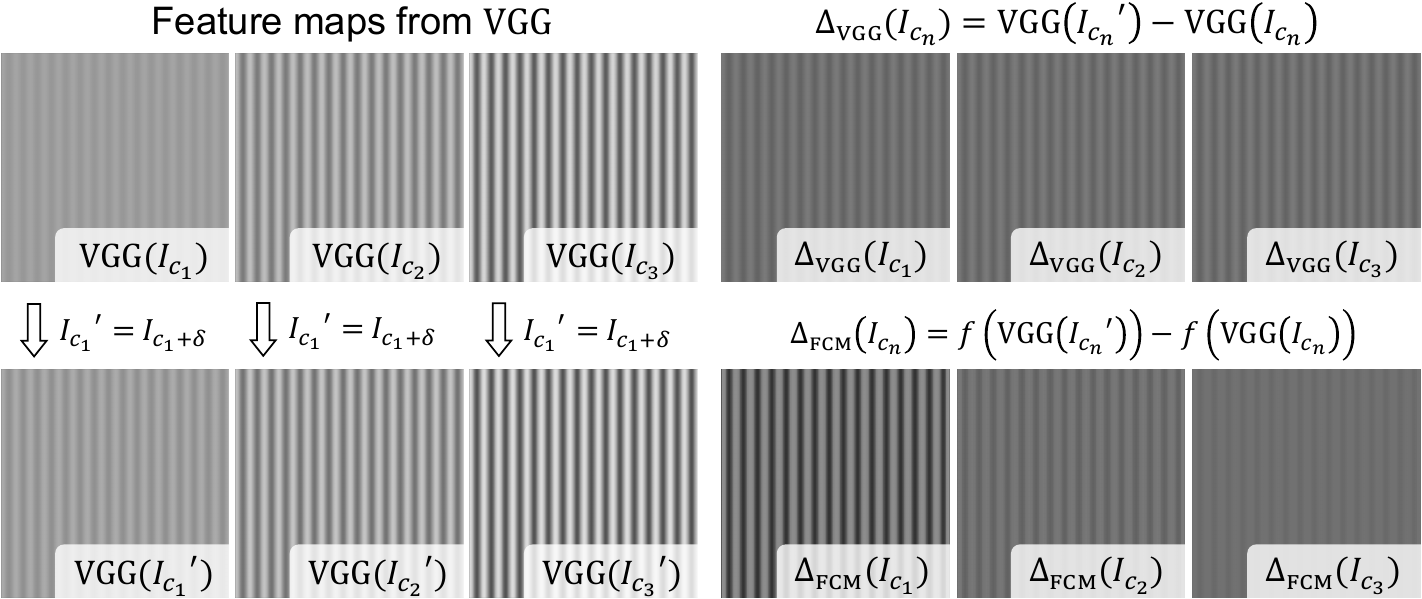}
	\caption{\label{fig:loss_comparison}	
		Left: VGG feature maps (1st layer, 2nd channel) for three input sinusoids of increasing image contrast ($c_1 < c_2 < c_3$) (upper row) that is distorted by further increasing their amplitudes by the same factor $\delta$ (bottom row).
		Right: the differences $\Delta_{\scalebox{.9}{$\scriptscriptstyle \textrm{VGG}$}}$ between the VGG feature maps, which are computed for each sinusoid and its distorted version, show a weak dependence to the input sinusoid contrast (upper row), while the corresponding differences $\Delta_{\scalebox{.9}{$\scriptscriptstyle \textrm{FCM}$}}$ resulting from our feature contrast masking model stronger penalize the distortion for smaller contrast sinusoids. 
	}
\end{figure}

To further illustrate the behavior of our loss, In Fig.~\ref{fig:loss_comparison} we consider simple sinusoid patterns with three different contrasts ($c_1$, $c_2$, and $c_3$). The corresponding feature maps of the VGG network are shown in the top-left image row. We distort each sinusoid by increasing their respective amplitudes by the same factor $\delta$ and their corresponding feature maps are shown in the bottom-left row. 

We compare the feature map difference $\Delta_{\scalebox{.9}{$\scriptscriptstyle \textrm{VGG}$}}$ used in the VGG loss $\mathcal{L}_{\scalebox{.9}{$\scriptscriptstyle \textrm{VGG}$}}$ \cite{johnson2016perceptual} (top-right row) and the corresponding $\Delta_{\scalebox{.9}{$\scriptscriptstyle \textrm{FCM}$}}$ used in our FCM loss $\mathcal{L}_{\scalebox{.9}{$\scriptscriptstyle \textrm{FCM}$}}$ (bottom-right row). As can be seen,  $\Delta_{\scalebox{.9}{$\scriptscriptstyle \textrm{FCM}$}}$ gives highest penalty when the distortion $\delta$ is added to the lowest contrast pattern. 
This forces our tone mapping, driven by $\mathcal{L}_{\scalebox{.9}{$\scriptscriptstyle \textrm{FCM}$}}$, to reproduce image details in low-contrast areas. 
The VGG loss $\mathcal{L}_{\scalebox{.9}{$\scriptscriptstyle \textrm{VGG}$}}$ remains similar irrespectively on the input sinusoid contrast that puts equal pressure on the tone mapping network to reproduce image details for large contrast regions that cannot be perceived, and low-contrast regions where they are clearly visible.

\section{Results and Ablation Study}
\label{experiment}
In this section we first describe the implementation details of our approach. Then, we provide objective and subjective comparisons including both traditional methods and state-of-the-art learning-based approaches. Finally, we perform an ablation study showing how each of the components of our approach contributes to achieving the final quality of our results. 

\subsection{Implementation}
\label{implementation}
We adopt an online training strategy and train a model for each HDR image at test time. Our model is implemented on TensorFlow and the results reported in the paper are computed with a RTX 8000 GPU. We use the Adam optimizer with an initial learning rate of 2 $\times$ $10^{-4}$ and an exponential decay factor of 0.9 every ten epochs. The training converges after 400 epochs, which translates into around 583 $\pm$ 6.62 seconds, for an image resolution of 768 $\times$ 384. 
We fix a single set of parameters for all our experiments and results. As discussed in the previous section, we set $\mathcal{P} =$ 13$\times$13,  $\alpha_{I_{\mu}}=0.5$, and $\alpha_{I_{\scalebox{.9}{$\scriptscriptstyle \textrm{TM}$}}}=1.0$. We compute our loss function based on the first three layers of VGG. Please refer to the supplemental for experimental exploration of these parameters. 

\subsection{Results and comparisons}

We include in our comparisons thirteen tone mapping operators including nine traditional methods which for simplicity we refer to as: Mantiuk~\cite{mantiuk2006perceptual}, Shan~\cite{shan2009globally}, Durand~\cite{durand2002fast}, Drago~\cite{drago2003adaptive}, Mertens~\cite{mertens2007exposure}, Reinhard~\cite{reinhard2002photographic}, Liang~\cite{liang2018hybrid}, Shibata~\cite{shibata2016gradient}, and Li~\cite{li2018clustering};
and four recent learning-based methods: Guo~\cite{guo2021deep}, Zhang~\cite{9454333}, DeepTMO~\cite{rana2019deep} and TMO-Net~\cite{panetta2021tmo}. 
We use the publicly available implementations of these methods or if not available, their implementation in HDRToolBox~\cite{banterle2017advanced}. For the case of DeepTMO~\cite{rana2019deep} and TMO-Net~\cite{panetta2021tmo}, we were not able to access their implementations, therefore we directly use the results provided in their works for comparisons. 

We use a large test set of 275 images gathered from the Fairchild dataset~\cite{fairchild2007hdr}, Poly Haven$\footnote{\url{https://polyhaven.com/hdris}}$, the Laval Indoor HDR Database~\cite{gardner2017learning}, and the LVZ-HDR benchmark dataset~\cite{panetta2021tmo}, which cover various indoors, outdoors, bright and dark scenes.  

\subsubsection{Objective evaluation}
\label{objective comparisons}

\begin{table*}
	\caption{\label{table: evaluation results}
		Mean and standard deviation for TMQI (including $\mathrm{TMQI_{S}}$ and $\mathrm{TMQI_{N}}$), BTMQI and BRISQUE computed for the 275 images in our test set.}
	\centering
	\begin{tabular}{|c | c | c | c | c | c | } 
		\hline
		Methods & TMQI ($\uparrow$) &$\mathrm{TMQI_{S}}$ ($\uparrow$) & $\mathrm{TMQI_{N}}$ ($\uparrow$)  & BTMQI ($\downarrow$) & BRISQUE ($\downarrow$)\\ [0.5ex] 
		\hline
		Ours & \textbf{0.9248 $\pm$ 0.0432} &\textbf{0.8938 $\pm$ 0.0582} &  \textbf{0.7062 $\pm$ 0.2369} & \textbf{2.9065 $\pm$ 1.0481} & \textbf{21.2208 $\pm$ 8.5981} \\ 
		\hline
		Guo \cite{guo2021deep} & 0.8883 $\pm$ 0.0354  & 0.8166 $\pm$ 0.0762 & 0.5975 $\pm$ 0.2002  &3.7110 $\pm$ 1.0239  &  23.9528 $\pm$ 8.2566 \\         
		\hline
		Zhang \cite{9454333} & 0.8767 $\pm$ 0.0600 &0.8343 $\pm$ 0.0872 & 0.5118 $\pm$ 0.2686 & 3.7440 $\pm$ 1.4209 & 22.0520 $\pm$ 8.9580 \\ 
		\hline
		Liang \cite{liang2018hybrid} & 0.8964 $\pm$ 0.0490 &0.8534 $\pm$  0.0737 & 0.5194 $\pm$ 0.2605 & 3.5333 $\pm$ 1.0915  & 25.1433 $\pm$ 8.5510 \\
		\hline
		Shibata \cite{shibata2016gradient} & 0.7689 $\pm$ 0.0506 &0.7498 $\pm$ 0.0858 &  0.1203 $\pm$ 0.1719 & 4.3470  $\pm$ 0.7469 & 32.2273 $\pm$ 8.9383\\
		\hline
		Li \cite{li2018clustering} & 0.8480 $\pm$ 0.0612 &0.8301 $\pm$ 0.0743 & 0.3716 $\pm$ 0.2939 & 4.3670 $\pm$ 1.1621 & 24.0348 $\pm$ 8.5411\\
		\hline
		Shan \cite{shan2009globally} &  0.8301 $\pm$ 0.0732  & 0.7458 $\pm$ 0.1448 & 0.4174 $\pm$ 0.2792 & 4.0685 $\pm$ 0.9839  & 22.8230 $\pm$ 8.5913\\ 
		\hline
		Durand \cite{durand2002fast} & 0.8719 $\pm$ 0.0669 &0.8375 $\pm$ 0.0989 & 0.4824 $\pm$ 0.2685 & 3.6537 $\pm$ 1.1016 & 22.0285 $\pm$ 8.4576\\
		\hline
		Drago \cite{drago2003adaptive} & 0.8794 $\pm$ 0.0537 & 0.8600 $\pm$0.0803  & 0.4840 $\pm$ 0.2558 & 4.0213 $\pm$ 1.2733 &  22.6600 $\pm$  8.2405\\
		\hline
		Mertens \cite{mertens2007exposure} & 0.8425 $\pm$ 0.0717 &0.8403 $\pm$ 0.0923 & 0.3373 $\pm$  0.2848 & 4.8981 $\pm$ 1.6026 & 23.2085 $\pm$ 8.8589\\
		\hline
		Reinhard \cite{reinhard2002photographic} & 0.8506 $\pm$ 0.0533 & 0.8176 $\pm$ 0.0864  & 0.3903 $\pm$ 0.2347 & 4.2774 $\pm$ 1.4580 & 25.5317 $\pm$ 7.7396\\ 
		\hline
		Mantiuk \cite{mantiuk2006perceptual} & 0.8529 $\pm$ 0.0753 & 0.8903 $\pm$ 0.0849 & 0.3238 $\pm$ 0.3050 & 4.5339 $\pm$ 1.4086 & 21.2943 $\pm$ 8.8302 \\ [0.5ex] 
		\hline
	\end{tabular}
\end{table*}

\begin{table*}
	\caption{\label{table: deeptmo evaluation results}%
		Mean and standard deviation for TMQI (including $\mathrm{TMQI_{S}}$ and $\mathrm{TMQI_{N}}$), BTMQI and BRISQUE computed for the DeepTMO~\cite{rana2019deep} and TMO-net~\cite{panetta2021tmo} test sets. The former contains 100 images from the Fairchild dataset~\cite{fairchild2007hdr} while the latter contains 457 captured images from their own dataset.
	}
	\centering
	\begin{tabular}{|c | c | c | c | c | c | } 
		\hline
		Methods & $\mathrm{TMQI_{Q}}$ ($\uparrow$) &$\mathrm{TMQI_{S}}$ ($\uparrow$) & $\mathrm{TMQI_{N}}$ ($\uparrow$)  & BTMQI ($\downarrow$) & BRISQUE ($\downarrow$)\\ 
		\hline
		Ours & \textbf{0.9106 $\pm$ 0.0511} &\textbf{0.8987 $\pm$ 0.0664}  & \textbf{0.6052 $\pm$ 0.2807}  & \textbf{3.3420 $\pm$ 1.0686}  &\textbf{19.5406 $\pm$ 9.1347}\\ 
		\hline 
		DeepTMO \cite{rana2019deep} & 0.9052 $\pm$ 0.0619 & 0.8810 $\pm$ 0.0717  &0.6015 $\pm$ 0.2679 & 3.4230 $\pm$ 1.1502 & 27.5489 $\pm$ 7.6865\\ 
		\hline \hline 		
		Ours & \textbf{0.9073 $\pm$ 0.0541} &\textbf{0.8939 $\pm$ 0.0551}  & \textbf{0.6020 $\pm$ 0.3126}  & \textbf{3.3069 $\pm$ 1.2266}  &\textbf{23.1010 $\pm$ 8.5232}\\ 
		\hline 
		TMO-Net \cite{panetta2021tmo} &0.8609 $\pm$ 0.0594 & 0.8066 $\pm$ 0.0825  &0.4723 $\pm$ 0.2871 & 3.9633 $\pm$ 1.2477 & 26.6078 $\pm$ 8.1895\\
		\hline
	\end{tabular}
\end{table*}

For objective comparisons we adopt as metrics the Tone Mapped Image Quality Index (TMQI)~\cite{yeganeh2012objective}, the Blind Tone Mapped Quality Index
(BTMQI)~\cite{gu2016blind} and the Blind/Referenceless Image Spatial Quality Evaluator (BRISQUE)~\cite{mittal2012no}. The former two metrics are widely used for evaluating tone mapping operators~\cite{guo2021deep, 9454333, liang2018hybrid}, while the latter is typically used as a blind metric for evaluating contrast~\cite{jiang2019nighttime, lim2017contrast, song2020enhancement}. We briefly discuss here these metrics, please refer to the supplemental for a more detailed description. 

TMQI is a full-reference tone mapping image quality metric, which consists of two main terms assessing the structural fidelity ($\mathrm{TMQI_{S}}$) and the naturalness ($\mathrm{TMQI_{N}}$) of the tone mapped image.
BTMQI is a no-reference tone mapping metric, which is composed of three terms accounting for entropy (richness of information), naturalness, and presence of structural details. 
BRISQUE is a well-known no-reference image quality metric based on natural scene statistics that quantifies the naturalness of an image and considers distortions such as noise, ringing, blur, or blocking artifacts. 

\begin{figure*}[h!]
	\centering
	\includegraphics[width=0.9\linewidth]{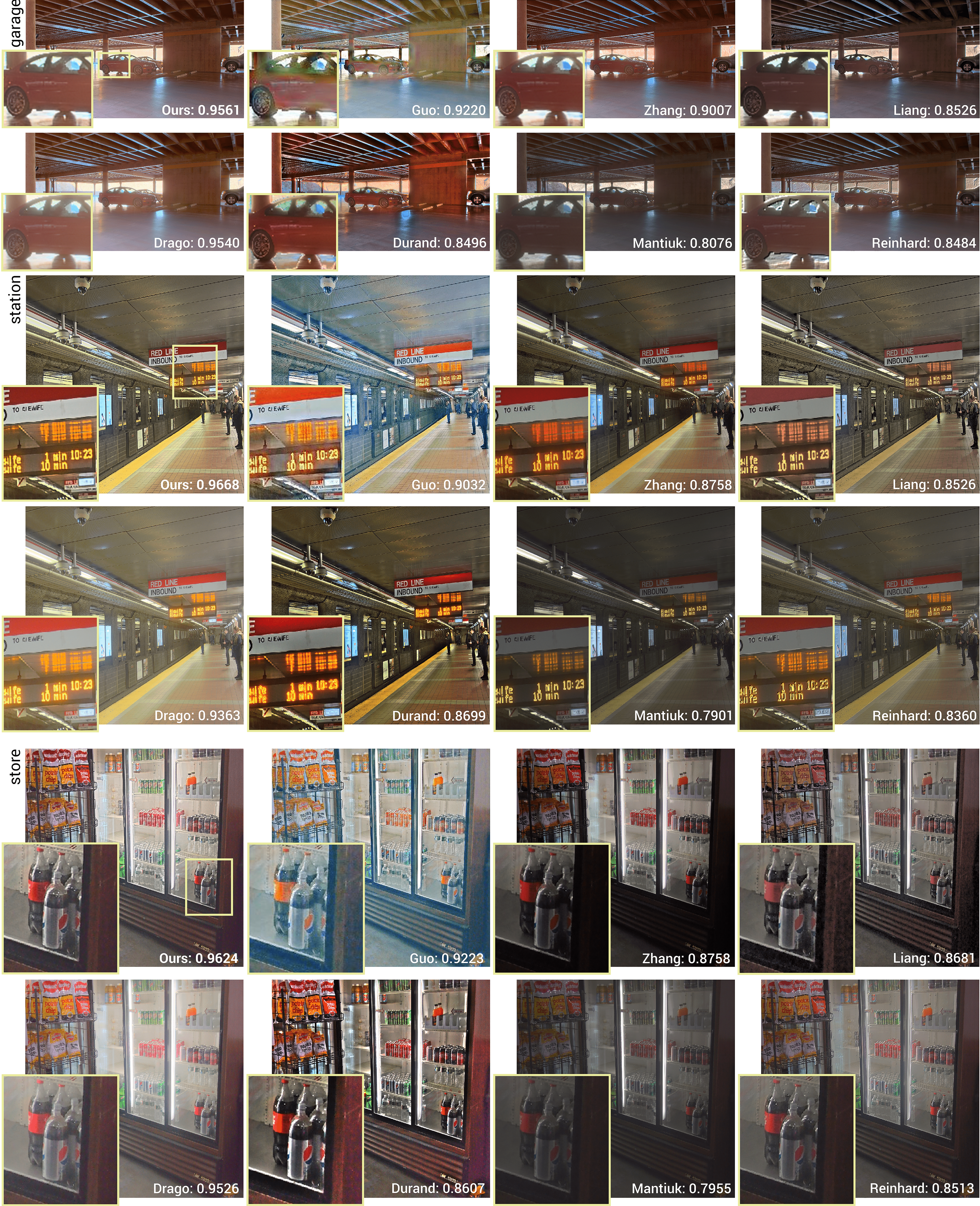}
	\caption{\label{fig:comparison}	
		Visual comparisons for the best performing methods and their TMQI scores. Overall, our results achieve good contrast reproduction while preserving the details (highlights of the bottles in \emph{store}, floor in \emph{station}) and avoiding visual artifacts (display sign in \emph{station}, car edges and windows in \emph{garage}). Please refer to the text for an in-depth discussion of the observed differences.}
\end{figure*}

\begin{figure*}[bth]
	\centering
	\includegraphics[width=\linewidth]{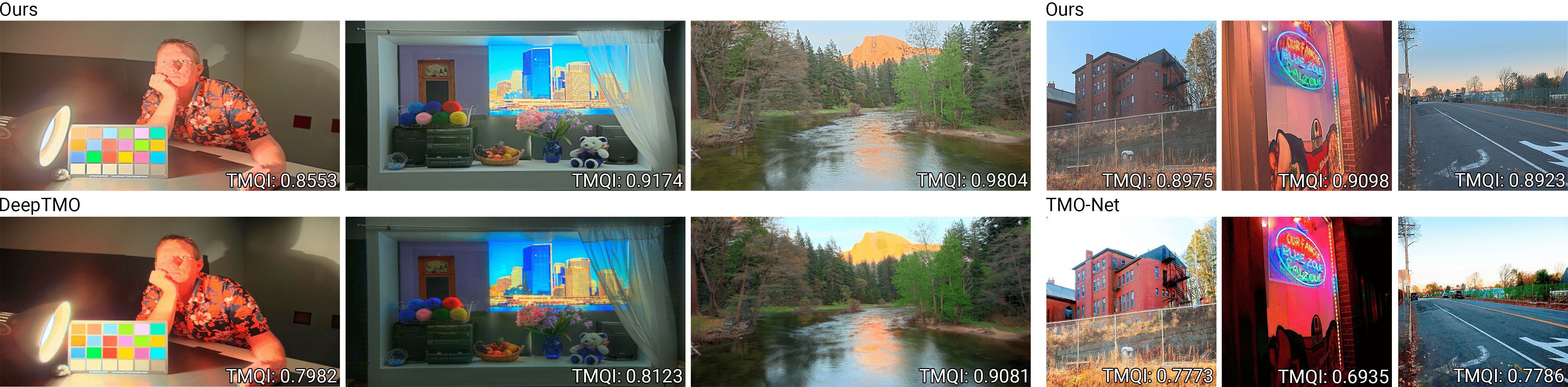}
	\caption{\label{fig: deeptmo-tmonet}	
		Visual comparisons for DeepTMO~\cite{rana2019deep} and TMO-Net~\cite{panetta2021tmo} on images from their respective test set. DeepTMO tends to produce over-enhanced contrast and saturated results that do not preserve natural image statistics. TMO-Net can not handle some challenging scenarios, producing saturated results in very bright regions and overly dark pixels in dark regions.}
\end{figure*}

We show in Table~\ref{table: evaluation results} the results of these objective metrics for eleven of the tested methods with our testing set, while Table~\ref{table: deeptmo evaluation results} shows the results for DeepTMO and TMO-Net, for which we use their provided test sets and results. We include examples of our results compared to the seven best performing methods in Fig.~\ref{fig:comparison}, and compared to DeepTMO and TMO-Net in Fig.~\ref{fig: deeptmo-tmonet}. We include more results and comparisons in our supplemental.

Our proposed approach outperforms previous methods in different aspects. We discuss now more in detail results for the best performing approaches.
In general, all approaches except DeepTMO yield low $\mathrm{TMQI_{N}}$ values, indicating that they fall short in preserving the naturalness of the image. For the case of DeepTMO, the low performance in the BRISQUE metric indicates that the tone mapped images do not preserve natural image statistics (Fig.~\ref{fig: deeptmo-tmonet}, first three columns). We can also see that TMO-Net additionally produces saturated results in the brightest regions of the image (Fig.~\ref{fig: deeptmo-tmonet}, last three columns).
Looking into Fig.~\ref{fig:comparison} we can observe that Guo over-enhances dark regions, sometimes resulting in heavy artifacts (\emph{garage} scene), while Zhang tends to produce overly dark results in regions with low brightness. This is in agreement with a relatively low score in structural fidelity $\mathrm{TMQI_{S}}$, indicating that the tone mapped images differ from the HDR in terms of conveyed structural information. 
While Liang performs well in terms of $\mathrm{TMQI_{S}}$, it tends to produce under-saturated results with lower contrast (e.g., \emph{garage} and \emph{store} scenes), which is in agreement with a relatively low performance in the BRISQUE metric.  
Drago and Durand also perform well in terms of $\mathrm{TMQI_{S}}$. However, we can see that Drago produces blurry results and fails to reproduce fine details, such as the floor tiling in the \emph{station} scene or the highlights of the bottles in the \emph{store} scene.
Durand presents good scores in terms of BRISQUE score which means the tone mapped images do align with natural image statistics, however in some cases it over-enhances contrast, producing artifacts such as those around the car windows in the \emph{garage} scene or those in the luminous numbers of the \emph{station} display sign. For these same regions, when the images are tone mapped with Reinhard, we can observe halos due to strong contrast edges. 
Finally, Mantiuk has the lowest $\mathrm{TMQI_{N}}$ of the methods included in Fig.~\ref{fig:comparison}. This method tends to produce very dark images with low contrast. 

Our method outperforms existing approaches for all tested metrics, exhibiting a good contrast reproduction while preserving the details present in the HDR images. Our results also produce natural images without visible artifacts.

\subsubsection{Subjective evaluation}
\label{subjective comparisons}

To further validate the performance of our approach we additionally performed a subjective study. We included the six best performing methods in terms of average TMQI score according to our previous objective evaluation, in particular: Guo~\cite{guo2021deep}, Zhang~\cite{9454333}, Liang~\cite{liang2018hybrid}, Drago~\cite{drago2003adaptive}, Durand~\cite{durand2002fast}, and ours. 
The study was approved by the \textit{<hidden for anonymity>} Ethical Board, and participants provided written consent for participating in the study. 
A total of twelve participants voluntarily took part in the study. 
We included fifteen scenes covering different scenarios and, for each scene, we showed the six tone mapped images at a random order. We asked the participants to rank them from 1 (preferred) to 6 (least preferred). The images were displayed in a DELL UltraSharp U2421E monitor (1920 $\times$ 1200 resolution, 60 HZ refresh rate).

\begin{figure}[htb]
	\centering
	\includegraphics[width=\linewidth]{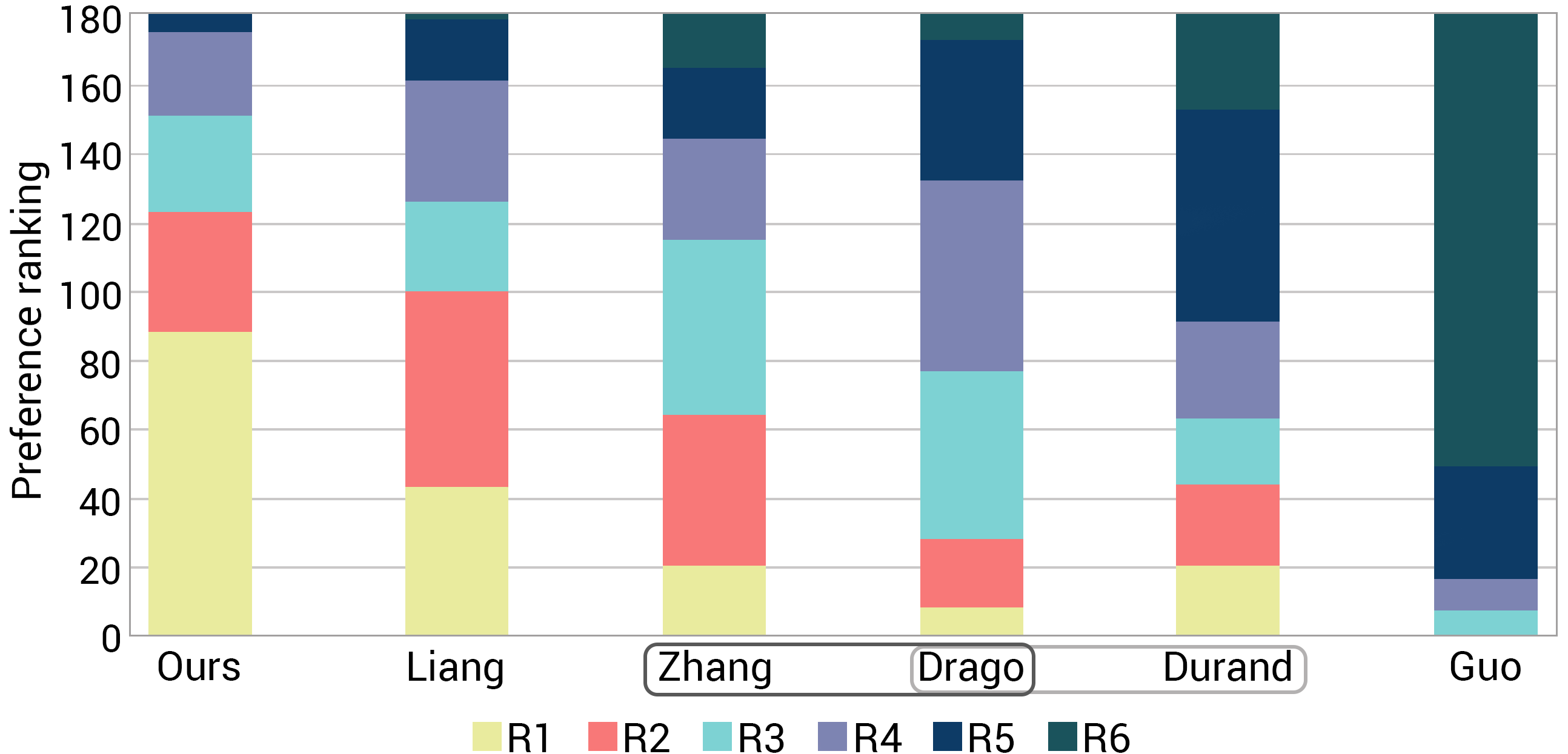}
	\caption{\label{fig: user_study}	
		Preference rankings for each method aggregated across the twelve participants and fifteen scenes. Different colors indicate the received rankings (from 1 to 6). Pairwise comparisons between methods reveal that preference rankings are significantly different, except those methods marked in the same set (gray squares), which are statistically indistinguishable.}
\end{figure}

We show in Fig.~\ref{fig: user_study} the preference rankings for each method, aggregated for all participants and scenes (refer to the supplemental for individual results for each scene). 
We use Kruskal-Wallis (non-parametric extension of ANOVA) for analyzing the rankings, since these do not follow a normal distribution~\cite{jarabo2014people, rubinstein2010comparative}. We then compute post-hocs using pairwise Kruskal-Wallis tests adjusted by Bonferroni correction for multiple test. Results reveal a statistically significant difference in the rankings for the different methods ($p < 0.001$), with our approach being ranked significantly higher than all others (refer to the supplemental for statistical tests for all pairwise comparisons). 

\subsection{Ablation Studies}
\label{ablation studies}

\begin{table*}
	\caption{\label{tabel: ablation study}%
		Mean and standard deviation for TMQI (including $\mathrm{TMQI_{S}}$ and $\mathrm{TMQI_{N}}$), BTMQI and BRISQUE computed for different variations of components in our pipeline.}
	\centering
	\begin{tabular}{| c |c | c | c | c| c | c| c|} 
		\hline
		Input & $I_{\scalebox{.9}{$\scriptscriptstyle \textrm{HDR}$}} \mapsto I_\mu$ & Loss  & TMQI ($\uparrow$) &$\mathrm{TMQI_{S}}$ ($\uparrow$) & $\mathrm{TMQI_{N}}$ ($\uparrow$) & BTMQI ($\downarrow$) & BRISQUE($\downarrow$) \\ [0.5ex] 
		\hline
		log & log    & $\mathcal{L}_{\scalebox{.9}{$\scriptscriptstyle \textrm{VGG}$}}$ & 0.5785 $\pm$ 0.1073  & 0.3759$\pm$ 0.1965 & 0.0066$\pm$ 0.0578  & 6.6324 $\pm$  0.5731 & 42.5287 $\pm$ 7.4740\\ 
		\hline
		MEF  & linear   & $\mathcal{L}_{\scalebox{.9}{$\scriptscriptstyle \textrm{FCM}$}}$  & 0.8776 $\pm$ 0.0610  & 0.8763$\pm$ 0.0966 & 0.4555 $\pm$ 0.2642  & 3.9174 $\pm$  1.4472 & 21.4104 $\pm$ 8.9001\\ 
		\hline
		MEF  & Ada $\mu$  & $\mathcal{L}_{\scalebox{.9}{$\scriptscriptstyle \textrm{VGG}$}}$ & 0.9178 $\pm$ 0.0462  &0.8913 $\pm$ 0.0607 & 0.6596 $\pm$ 0.2470 & 3.1934 $\pm$ 1.1220  & 22.1773 $\pm$ 8.6294\\ 
		\hline
		linear & Ada $\mu$ & $\mathcal{L}_{\scalebox{.9}{$\scriptscriptstyle \textrm{FCM}$}}$  & 0.8522 $\pm$ 0.0793 & 0.8166$\pm$ 0.1206 & 0.4173 $\pm$  0.2902 & 4.5342$\pm$ 1.6754 & 30.2186 $\pm$ 13.7742\\ 
		\hline
		log & Ada $\mu$  & $\mathcal{L}_{\scalebox{.9}{$\scriptscriptstyle \textrm{FCM}$}}$   & 0.8166 $\pm$ 0.0563 & 0.8921 $\pm$ 0.0774 & 0.0844 $\pm$  0.0741 & 5.0949 $\pm$ 1.1779 & 22.4499 $\pm$ 8.3301\\ 
		\hline
		MEF & Ada $\mu$ & $\mathcal{L}_{\scalebox{.9}{$\scriptscriptstyle \textrm{FCM}$}}$  & \textbf{0.9248 $\pm$ 0.0432} &\textbf{0.8938 $\pm$ 0.0582} &  \textbf{0.7062 $\pm$ 0.2369} & \textbf{2.9065 $\pm$ 1.0481} & \textbf{21.2208 $\pm$ 8.5981} \\ 
		\hline
	\end{tabular}
\end{table*}

\begin{figure*}[tbh]
	\centering
	\includegraphics[width=\linewidth]{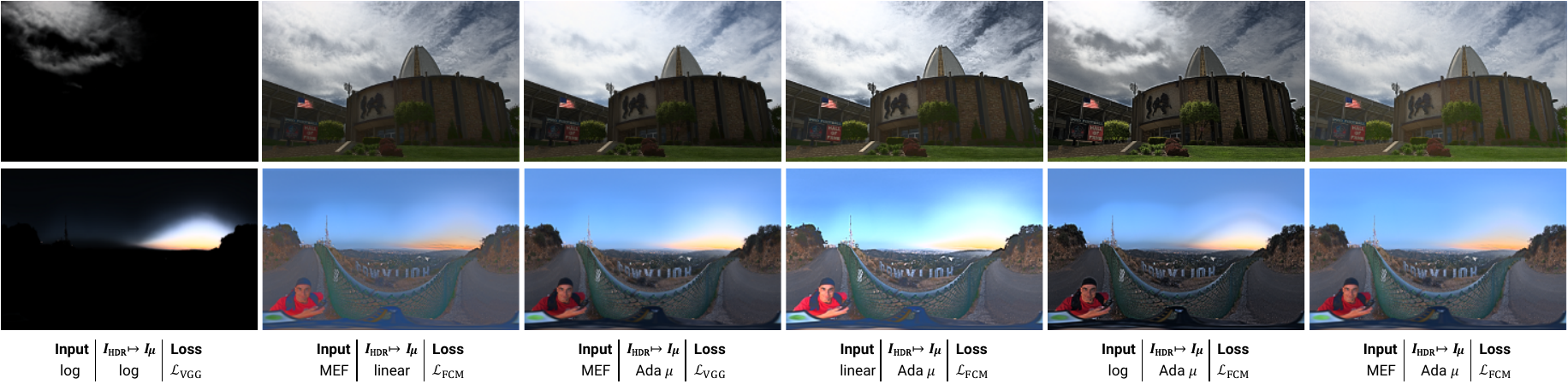}
	\caption{\label{fig: ablation study}	
		Example visualizations of our ablation study. Better contrast, brightness and detail reproduction is achieved with our full pipeline including multiple exposure fusion (\emph{MEF}), adaptive $\mu$-law compression (\emph{Ada $\mu$}), and $\mathcal{L}_{\scalebox{.9}{$\scriptscriptstyle \textrm{FCM}$}}$ loss. 
	}
\end{figure*}

In this section we evaluate the importance of each of the components in our method for achieving the final quality of the results. Table~\ref{tabel: ablation study} shows the results of the objective metrics for different combinations of (i) input: linear HDR~\cite{rana2019deep}, log HDR~\cite{zhang2019deep, su2021explorable} or our multiple exposure fusion (\emph{MEF}); (ii) HDR compression algorithm for computing the loss: linear, log or our adaptive $\mu$-law compression (\emph{Ada $\mu$}); and (iii) loss function: $\mathcal{L}_{\scalebox{.9}{$\scriptscriptstyle \textrm{VGG}$}}$ or our $\mathcal{L}_{\scalebox{.9}{$\scriptscriptstyle \textrm{FCM}$}}$. Figure \ref{fig: ablation study} shows the corresponding visual results. Please, refer to the supplemental for extended results on the ablation. 

In general, we can see that our loss $\mathcal{L}_{\scalebox{.9}{$\scriptscriptstyle \textrm{FCM}$}}$, which considers masking effects, plays an important role in emphasizing the local contrast, especially in large contrast regions, such as the clouds in the sky. 
Compared with our multiple exposure fusion (MEF), the logarithm input leads to overall darker results (brightness distortion), and the linear input can cause overexposure with missing information in the highlight regions.
Finally, we can see Ada $\mu$-law compression is important for overall image contrast, when linear or logarithmic transformations are applied instead, the resulting images are flatter in terms of contrast. 

\section{Conclusions}

In this work we propose an image-specific self-supervised tone mapping approach that leads to consistent high-quality results for a large variety of HDR scenes.
Previous learning-based approaches present two main limitations: (i) the variety of HDR content they can represent and adequately tone map is limited by the images used during training, and (ii) most of these approaches are supervised, i.e., they need HDR-LDR image pairs for training. These LDR images are obtained either from tone mapped results from previous methods, or manually tone mapped images. Therefore, the quality of the trained tone mapping model is limited to that of the results selected for training.  
In contrast, our network directly learns to represent the input HDR image and the tone mapping process is guided by a novel feature contrast masking model that allows for representing important image contrast perception characteristics of the Human Visual System directly in the feature space.
This in turn enables to derive a powerful and perceptually meaningful loss that uses such feature contrast masking to guide the tone mapping operation.
The loss gives the network more freedom for compressing higher contrast while enhancing weak contrast, as it is often desirable for high quality HDR scene reproduction and an overall pleasant appearance, all in the context of local image content as modeled by neighborhood masking. 

\begin{figure}[thb]
	\centering
	\includegraphics[width=\linewidth]{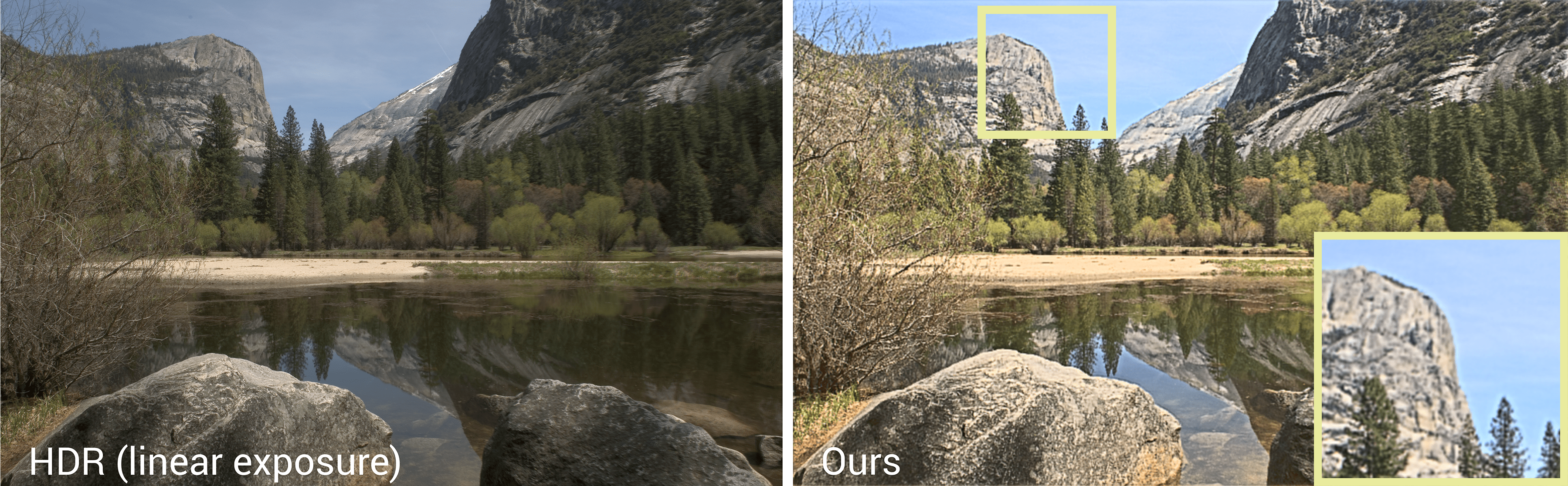}
	\hfill
	\caption{\label{fig:limitations}%
		Example exposure of the original HDR image (left) and our tone mapped result (right). The inset shows a failure case in which a soft halo appears around the edge of the mountain. 
	}
\end{figure}

\paragraph*{Limitations and future work} In some very rare cases our method may produce soft halos at high contrast edges as shown in Fig.~\ref{fig:limitations}.
In future work we would like to experiment with edge stopping filters while deriving feature contrast and neighborhood masking for avoiding this issue. Nevertheless, these soft artifacts are not present in most of our results, and whenever present, they are not obviously visible as confirmed by both objective and subjective evaluations.
As discussed in Trentacoste et al.~\cite{trentacoste2012unsharp}, unsharp masking or weak counter-shading effects, similar to these soft halos, may be an effective way of enhancing perceived image contrast due to the Cornsweet illusion and are often employed for image enhancement.
In future work we would also like to accelerate our tone mapping processing, which is the key limitation of our technique.
We would also like to investigate the utility of our adaptive $\mu$-law compression for other learning-based applications that involve HDR content.
Finally, another interesting avenue for future work would be employing our feature contrast masking model for other tasks such as image style transfer, where contrast characteristics in the source image should be conveyed to the target image.
%
%

\bibliographystyle{eg-alpha-doi} 
\bibliography{egbib}      

\newcommand{\etalchar}[1]{$^{#1}$}
\begin{thebibliography}{\uppercase{TAKW{\etalchar{*}}19}}

\bibitem[AJP92]{ahumada1992luminance}
\textsc{Ahumada~Jr A.~J., Peterson H.~A.}:
\newblock Luminance-model-based dct quantization for color image compression.
\newblock In \emph{Human vision, visual processing, and digital display III}
  (1992), vol.~1666, International Society for Optics and Photonics,
  pp.~365--374.

\bibitem[ANSAM21]{andersson2021visualizing}
\textsc{Andersson P., Nilsson J., Shirley P., Akenine-M{\"o}ller T.}:
\newblock Visualizing errors in rendered high dynamic range images.

\bibitem[BADC17]{banterle2017advanced}
\textsc{Banterle F., Artusi A., Debattista K., Chalmers A.}:
\newblock \emph{Advanced high dynamic range imaging}.
\newblock AK Peters/CRC Press, 2017.

\bibitem[BM98]{bolin1998perceptually}
\textsc{Bolin M.~R., Meyer G.~W.}:
\newblock A perceptually based adaptive sampling algorithm.
\newblock In \emph{Proceedings of the 25th annual conference on Computer
  graphics and interactive techniques} (1998), pp.~299--309.

\bibitem[{\v{C}}WNA08]{vcadik2008evaluation}
\textsc{{\v{C}}ad{\'\i}k M., Wimmer M., Neumann L., Artusi A.}:
\newblock Evaluation of hdr tone mapping methods using essential perceptual
  attributes.
\newblock \emph{Computers \& Graphics 32}, 3 (2008), 330--349.

\bibitem[Dal92]{daly1992visible}
\textsc{Daly S.~J.}:
\newblock Visible differences predictor: an algorithm for the assessment of
  image fidelity.
\newblock In \emph{Human Vision, Visual Processing, and Digital Display III}
  (1992), vol.~1666, International Society for Optics and Photonics, pp.~2--15.

\bibitem[DD02]{durand2002fast}
\textsc{Durand F., Dorsey J.}:
\newblock Fast bilateral filtering for the display of high-dynamic-range
  images.
\newblock In \emph{Proceedings of the 29th annual conference on Computer
  graphics and interactive techniques} (2002), pp.~257--266.

\bibitem[DER{\etalchar{*}}10]{didyk2010apparent}
\textsc{Didyk P., Eisemann E., Ritschel T., Myszkowski K., Seidel H.-P.}:
\newblock Apparent display resolution enhancement for moving images.
\newblock In \emph{ACM SIGGRAPH 2010 papers}. 2010, pp.~1--8.

\bibitem[DMAC03]{drago2003adaptive}
\textsc{Drago F., Myszkowski K., Annen T., Chiba N.}:
\newblock Adaptive logarithmic mapping for displaying high contrast scenes.
\newblock In \emph{Computer graphics forum} (2003), vol.~22, Wiley Online
  Library, pp.~419--426.

\bibitem[DZLL00]{daly2000visual}
\textsc{Daly S.~J., Zeng W., Li J., Lei S.}:
\newblock Visual masking in wavelet compression for jpeg-2000.
\newblock In \emph{Image and Video Communications and Processing 2000} (2000),
  vol.~3974, International Society for Optics and Photonics, pp.~66--80.

\bibitem[EKD{\etalchar{*}}17]{eilertsen2017hdr}
\textsc{Eilertsen G., Kronander J., Denes G., Mantiuk R.~K., Unger J.}:
\newblock Hdr image reconstruction from a single exposure using deep cnns.
\newblock \emph{ACM transactions on graphics (TOG) 36}, 6 (2017), 1--15.

\bibitem[EKM17]{endoSA2017}
\textsc{Endo Y., Kanamori Y., Mitani J.}:
\newblock Deep reverse tone mapping.
\newblock \emph{ACM Transactions on Graphics (Proc. of SIGGRAPH ASIA 2017) 36},
  6 (Nov. 2017).

\bibitem[Fai07]{fairchild2007hdr}
\textsc{Fairchild M.~D.}:
\newblock The hdr photographic survey.
\newblock In \emph{Color and imaging conference} (2007), vol.~2007, Society for
  Imaging Science and Technology, pp.~233--238.

\bibitem[FLW02]{fattal2002gradient}
\textsc{Fattal R., Lischinski D., Werman M.}:
\newblock Gradient domain high dynamic range compression.
\newblock In \emph{Proceedings of the 29th annual conference on Computer
  graphics and interactive techniques} (2002), pp.~249--256.

\bibitem[Fol94]{foley1994human}
\textsc{Foley J.~M.}:
\newblock Human luminance pattern-vision mechanisms: masking experiments
  require a new model.
\newblock \emph{JOSA A 11}, 6 (1994), 1710--1719.

\bibitem[GJ21]{guo2021deep}
\textsc{Guo C., Jiang X.}:
\newblock Deep tone-mapping operator using image quality assessment inspired
  semi-supervised learning.
\newblock \emph{IEEE Access 9} (2021), 73873--73889.

\bibitem[GSY{\etalchar{*}}17]{gardner2017learning}
\textsc{Gardner M.-A., Sunkavalli K., Yumer E., Shen X., Gambaretto E.,
  Gagn{\'e} C., Lalonde J.-F.}:
\newblock Learning to predict indoor illumination from a single image.
\newblock \emph{arXiv preprint arXiv:1704.00090} (2017).

\bibitem[GWZ{\etalchar{*}}16]{gu2016blind}
\textsc{Gu K., Wang S., Zhai G., Ma S., Yang X., Lin W., Zhang W., Gao W.}:
\newblock Blind quality assessment of tone-mapped images via analysis of
  information, naturalness, and structure.
\newblock \emph{IEEE Transactions on Multimedia 18}, 3 (2016), 432--443.

\bibitem[HDQ17]{hou2017deep}
\textsc{Hou X., Duan J., Qiu G.}:
\newblock Deep feature consistent deep image transformations: Downscaling,
  decolorization and hdr tone mapping.
\newblock \emph{arXiv preprint arXiv:1707.09482} (2017).

\bibitem[HZRS16]{he2016deep}
\textsc{He K., Zhang X., Ren S., Sun J.}:
\newblock Deep residual learning for image recognition.
\newblock In \emph{Proceedings of the IEEE conference on computer vision and
  pattern recognition} (2016), pp.~770--778.

\bibitem[JAFF16]{johnson2016perceptual}
\textsc{Johnson J., Alahi A., Fei-Fei L.}:
\newblock Perceptual losses for real-time style transfer and super-resolution.
\newblock In \emph{European conference on computer vision} (2016), Springer,
  pp.~694--711.

\bibitem[JH93]{jack1993tone}
\textsc{Jack T., Holly R.}:
\newblock Tone reproduction for realistic images.
\newblock \emph{IEEE Computer Graphics and Applications 13}, 6 (1993), 42--48.

\bibitem[JKX{\etalchar{*}}11]{jinno2011mu}
\textsc{Jinno T., Kaida H., Xue X., Adami N., Okuda M.}:
\newblock $\mu$-law based hdr coding and its error analysis.
\newblock \emph{IEICE transactions on fundamentals of electronics,
  communications and computer sciences 94}, 3 (2011), 972--978.

\bibitem[JMB{\etalchar{*}}14]{jarabo2014people}
\textsc{Jarabo A., Masia B., Bousseau A., Pellacini F., Gutierrez D.}:
\newblock How do people edit light fields.
\newblock \emph{ACM Trans. Graph 33}, 4 (2014), 4.

\bibitem[JYL19]{jiang2019nighttime}
\textsc{Jiang X., Yao H., Liu D.}:
\newblock Nighttime image enhancement based on image decomposition.
\newblock \emph{Signal, Image and Video Processing 13}, 1 (2019), 189--197.

\bibitem[KW96]{kingdom1996contrast}
\textsc{Kingdom F.~A., Whittle P.}:
\newblock Contrast discrimination at high contrasts reveals the influence of
  local light adaptation on contrast processing.
\newblock \emph{Vision research 36}, 6 (1996), 817--829.

\bibitem[LCTS05]{ledda2005evaluation}
\textsc{Ledda P., Chalmers A., Troscianko T., Seetzen H.}:
\newblock Evaluation of tone mapping operators using a high dynamic range
  display.
\newblock \emph{ACM Transactions on Graphics (TOG) 24}, 3 (2005), 640--648.

\bibitem[LF80]{legge1980contrast}
\textsc{Legge G.~E., Foley J.~M.}:
\newblock Contrast masking in human vision.
\newblock \emph{Josa 70}, 12 (1980), 1458--1471.

\bibitem[LHLK17]{lim2017contrast}
\textsc{Lim J., Heo M., Lee C., Kim C.-S.}:
\newblock Contrast enhancement of noisy low-light images based on
  structure-texture-noise decomposition.
\newblock \emph{Journal of Visual Communication and Image Representation 45}
  (2017), 107--121.

\bibitem[LJZ18]{li2018clustering}
\textsc{Li H., Jia X., Zhang L.}:
\newblock Clustering based content and color adaptive tone mapping.
\newblock \emph{Computer Vision and Image Understanding 168} (2018), 37--49.

\bibitem[LRP97]{larson1997visibility}
\textsc{Larson G.~W., Rushmeier H., Piatko C.}:
\newblock A visibility matching tone reproduction operator for high dynamic
  range scenes.
\newblock \emph{IEEE Transactions on Visualization and Computer Graphics 3}, 4
  (1997), 291--306.

\bibitem[Lub95]{lubin1995visual}
\textsc{Lubin J.}:
\newblock A visual discrimination model for imaging system design and
  evaluation.
\newblock In \emph{Vision Models for Target Detection and Recognition: In
  Memory of Arthur Menendez}. World Scientific, 1995, pp.~245--283.

\bibitem[LXZ{\etalchar{*}}18]{liang2018hybrid}
\textsc{Liang Z., Xu J., Zhang D., Cao Z., Zhang L.}:
\newblock A hybrid l1-l0 layer decomposition model for tone mapping.
\newblock In \emph{Proceedings of the IEEE conference on computer vision and
  pattern recognition} (2018), pp.~4758--4766.

\bibitem[MDC{\etalchar{*}}21]{mantiuk2021fovvideovdp}
\textsc{Mantiuk R.~K., Denes G., Chapiro A., Kaplanyan A., Rufo G., Bachy R.,
  Lian T., Patney A.}:
\newblock Fovvideovdp: A visible difference predictor for wide field-of-view
  video.
\newblock \emph{ACM Transactions on Graphics (TOG) 40}, 4 (2021), 1--19.

\bibitem[MDK08]{mantiuk2008display}
\textsc{Mantiuk R., Daly S., Kerofsky L.}:
\newblock Display adaptive tone mapping.
\newblock In \emph{ACM SIGGRAPH 2008 papers}. 2008, pp.~1--10.

\bibitem[MKVR07]{mertens2007exposure}
\textsc{Mertens T., Kautz J., Van~Reeth F.}:
\newblock Exposure fusion.
\newblock In \emph{15th Pacific Conference on Computer Graphics and
  Applications (PG'07)} (2007), IEEE, pp.~382--390.

\bibitem[MMB12]{mittal2012no}
\textsc{Mittal A., Moorthy A.~K., Bovik A.~C.}:
\newblock No-reference image quality assessment in the spatial domain.
\newblock \emph{IEEE Transactions on image processing 21}, 12 (2012),
  4695--4708.

\bibitem[MMS06]{mantiuk2006perceptual}
\textsc{Mantiuk R., Myszkowski K., Seidel H.-P.}:
\newblock A perceptual framework for contrast processing of high dynamic range
  images.
\newblock \emph{ACM Transactions on Applied Perception (TAP) 3}, 3 (2006),
  286--308.

\bibitem[MMS15]{mantiuk2015ency}
\textsc{Mantiuk R., Myszkowski K., Seidel H.-P.}:
\newblock \emph{High dynamic range imaging}.
\newblock Wiley Encyclopedia of Electrical and Electronics Engineering, 2015.

\bibitem[NH10]{nair2010rectified}
\textsc{Nair V., Hinton G.~E.}:
\newblock Rectified linear units improve restricted boltzmann machines.
\newblock In \emph{Icml} (2010).

\bibitem[Pal99]{palmer1999vision}
\textsc{Palmer S.~E.}:
\newblock \emph{Vision science: Photons to phenomenology}.
\newblock MIT press, 1999.

\bibitem[Pel90]{peli1990contrast}
\textsc{Peli E.}:
\newblock Contrast in complex images.
\newblock \emph{JOSA A 7}, 10 (1990), 2032--2040.

\bibitem[PKO{\etalchar{*}}21]{panetta2021tmo}
\textsc{Panetta K., Kezebou L., Oludare V., Agaian S., Xia Z.}:
\newblock Tmo-net: A parameter-free tone mapping operator using generative
  adversarial network, and performance benchmarking on large scale hdr dataset.
\newblock \emph{IEEE Access 9} (2021), 39500--39517.

\bibitem[PSR17]{patel2017generative}
\textsc{Patel V.~A., Shah P., Raman S.}:
\newblock A generative adversarial network for tone mapping hdr images.
\newblock In \emph{National Conference on Computer Vision, Pattern Recognition,
  Image Processing, and Graphics} (2017), Springer, pp.~220--231.

\bibitem[RC09]{raman2009bilateral}
\textsc{Raman S., Chaudhuri S.}:
\newblock Bilateral filter based compositing for variable exposure photography.
\newblock In \emph{Eurographics (short papers)} (2009), pp.~1--4.

\bibitem[RGSS10]{rubinstein2010comparative}
\textsc{Rubinstein M., Gutierrez D., Sorkine O., Shamir A.}:
\newblock A comparative study of image retargeting.
\newblock In \emph{ACM SIGGRAPH Asia 2010 papers}. 2010, pp.~1--10.

\bibitem[RHD{\etalchar{*}}10]{reinhard2010high}
\textsc{Reinhard E., Heidrich W., Debevec P., Pattanaik S., Ward G., Myszkowski
  K.}:
\newblock \emph{High dynamic range imaging: acquisition, display, and
  image-based lighting}.
\newblock Morgan Kaufmann, 2010.

\bibitem[Rob66]{robson1966spatial}
\textsc{Robson J.~G.}:
\newblock Spatial and temporal contrast-sensitivity functions of the visual
  system.
\newblock \emph{Josa 56}, 8 (1966), 1141--1142.

\bibitem[RPG99]{ramasubramanian1999perceptually}
\textsc{Ramasubramanian M., Pattanaik S.~N., Greenberg D.~P.}:
\newblock A perceptually based physical error metric for realistic image
  synthesis.
\newblock In \emph{Proceedings of the 26th annual conference on Computer
  graphics and interactive techniques} (1999), pp.~73--82.

\bibitem[RSSF02]{reinhard2002photographic}
\textsc{Reinhard E., Stark M., Shirley P., Ferwerda J.}:
\newblock Photographic tone reproduction for digital images.
\newblock In \emph{Proceedings of the 29th annual conference on Computer
  graphics and interactive techniques} (2002), pp.~267--276.

\bibitem[RSV{\etalchar{*}}19]{rana2019deep}
\textsc{Rana A., Singh P., Valenzise G., Dufaux F., Komodakis N., Smolic A.}:
\newblock Deep tone mapping operator for high dynamic range images.
\newblock \emph{IEEE Transactions on Image Processing 29} (2019), 1285--1298.

\bibitem[SJB09]{shan2009globally}
\textsc{Shan Q., Jia J., Brown M.~S.}:
\newblock Globally optimized linear windowed tone mapping.
\newblock \emph{IEEE transactions on visualization and computer graphics 16}, 4
  (2009), 663--675.

\bibitem[STKK20]{Marcel:2020:LDRHDR}
\textsc{Santos M.~S., Tsang R., Khademi~Kalantari N.}:
\newblock Single image hdr reconstruction using a cnn with masked features and
  perceptual loss.
\newblock \emph{ACM Transactions on Graphics 39}, 4 (7 2020).
\newblock \href {https://doi.org/10.1145/3386569.3392403}
  {\path{doi:10.1145/3386569.3392403}}.

\bibitem[STO16]{shibata2016gradient}
\textsc{Shibata T., Tanaka M., Okutomi M.}:
\newblock Gradient-domain image reconstruction framework with intensity-range
  and base-structure constraints.
\newblock In \emph{Proceedings of the IEEE conference on computer vision and
  pattern recognition} (2016), pp.~2745--2753.

\bibitem[SWH{\etalchar{*}}20]{song2020enhancement}
\textsc{Song W., Wang Y., Huang D., Liotta A., Perra C.}:
\newblock Enhancement of underwater images with statistical model of background
  light and optimization of transmission map.
\newblock \emph{IEEE Transactions on Broadcasting 66}, 1 (2020), 153--169.

\bibitem[SWL{\etalchar{*}}21]{su2021explorable}
\textsc{Su C.-C., Wang R., Lin H.-J., Liu Y.-L., Chen C.-P., Chang Y.-L., Pei
  S.-C.}:
\newblock Explorable tone mapping operators.
\newblock In \emph{2020 25th International Conference on Pattern Recognition
  (ICPR)} (2021), IEEE, pp.~10320--10326.

\bibitem[SZ14]{simonyan2014very}
\textsc{Simonyan K., Zisserman A.}:
\newblock Very deep convolutional networks for large-scale image recognition.
\newblock \emph{arXiv preprint arXiv:1409.1556} (2014).

\bibitem[TAKW{\etalchar{*}}19]{tursun2019luminance}
\textsc{Tursun O.~T., Arabadzhiyska-Koleva E., Wernikowski M., Mantiuk R.,
  Seidel H.-P., Myszkowski K., Didyk P.}:
\newblock Luminance-contrast-aware foveated rendering.
\newblock \emph{ACM Transactions on Graphics (TOG) 38}, 4 (2019), 1--14.

\bibitem[TMHD12]{trentacoste2012unsharp}
\textsc{Trentacoste M., Mantiuk R., Heidrich W., Dufrot F.}:
\newblock {Unsharp Masking, Countershading and Halos: Enhancements or
  Artifacts?}
\newblock In \emph{Proc. Eurographics} (2012), p.~to appear.

\bibitem[Wat89]{watson1989receptive}
\textsc{Watson A.~B.}:
\newblock Receptive fields and visual representations.
\newblock In \emph{Human Vision, Visual Processing, and Digital Display}
  (1989), vol.~1077, International Society for Optics and Photonics,
  pp.~190--197.

\bibitem[Wat93]{watson1993visually}
\textsc{Watson A.~B.}:
\newblock Visually optimal dct quantization matrices for individual images.
\newblock In \emph{[Proceedings] DCC93: Data Compression Conference} (1993),
  IEEE, pp.~178--187.

\bibitem[WS97]{watson1997model}
\textsc{Watson A.~B., Solomon J.~A.}:
\newblock Model of visual contrast gain control and pattern masking.
\newblock \emph{JOSA A 14}, 9 (1997), 2379--2391.

\bibitem[WXTT18]{wu2018hdr}
\textsc{Wu S., Xu J., Tai Y.-W., Tang C.-K.}:
\newblock Deep high dynamic range imaging with large foreground motions.
\newblock In \emph{The European Conference on Computer Vision (ECCV)} (2018).

\bibitem[YBMS05]{yoshida2005perceptual}
\textsc{Yoshida A., Blanz V., Myszkowski K., Seidel H.-P.}:
\newblock Perceptual evaluation of tone mapping operators with real-world
  scenes.
\newblock In \emph{Human Vision and Electronic Imaging X} (2005), vol.~5666,
  International Society for Optics and Photonics, pp.~192--203.

\bibitem[YJS{\etalchar{*}}21]{Yi2021_CSF}
\textsc{Yi S., Jeon D.~S., Serrano A., Jeong S.-Y., Kim H.-Y., Gutierrez D.,
  Kim M.~H.}:
\newblock Modeling surround-aware contrast sensitivity.
\newblock In \emph{Eurographics Symposium on Rendering} (2021), The
  Eurographics Association.
\newblock \href {https://doi.org/10.2312/sr.20211303}
  {\path{doi:10.2312/sr.20211303}}.

\bibitem[YW12]{yeganeh2012objective}
\textsc{Yeganeh H., Wang Z.}:
\newblock Objective quality assessment of tone-mapped images.
\newblock \emph{IEEE Transactions on Image Processing 22}, 2 (2012), 657--667.

\bibitem[YXS{\etalchar{*}}18]{yang2018image}
\textsc{Yang X., Xu K., Song Y., Zhang Q., Wei X., Lau R.~W.}:
\newblock Image correction via deep reciprocating hdr transformation.
\newblock In \emph{Proceedings of the IEEE Conference on Computer Vision and
  Pattern Recognition} (2018), pp.~1798--1807.

\bibitem[ZWZW19]{zhang2019deep}
\textsc{Zhang N., Wang C., Zhao Y., Wang R.}:
\newblock Deep tone mapping network in hsv color space.
\newblock In \emph{2019 IEEE Visual Communications and Image Processing (VCIP)}
  (2019), IEEE, pp.~1--4.

\bibitem[ZZP{\etalchar{*}}17]{zhu2017multimodal}
\textsc{Zhu J.-Y., Zhang R., Pathak D., Darrell T., Efros A.~A., Wang O.,
  Shechtman E.}:
\newblock Multimodal image-to-image translation by enforcing bi-cycle
  consistency.
\newblock In \emph{Advances in neural information processing systems} (2017),
  pp.~465--476.

\bibitem[ZZWW21]{9454333}
\textsc{Zhang N., Zhao Y., Wang C., Wang R.}:
\newblock A real-time semi-supervised deep tone mapping network.
\newblock \emph{IEEE Transactions on Multimedia} (2021), 1--1.
\newblock \href {https://doi.org/10.1109/TMM.2021.3089019}
  {\path{doi:10.1109/TMM.2021.3089019}}.

\end{thebibliography}


\newpage
\end{document}